\pdfoutput=1

\documentclass[11pt]{article}

\usepackage[table]{xcolor} %
\usepackage{acl}

\usepackage{booktabs, multirow} %
\usepackage{soul}%
\usepackage{changepage,threeparttable} %

\usepackage{times}
\usepackage{latexsym}

\usepackage[T1]{fontenc}

\usepackage[utf8]{inputenc}

\usepackage{microtype}
\usepackage{graphicx}
\usepackage{enumitem}

\title{CIF-Bench: A Chinese Instruction-Following Benchmark for Evaluating the Generalizability of Large Language Models}

\newcommand*\samethanks[1][\value{footnote}]{\footnotemark[#1]}

\newcommand{\maplogo}{
    \includegraphics[scale=0.03]{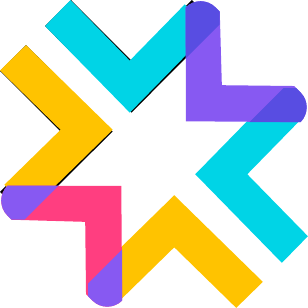}
}

\author{
{\scriptsize 
Yizhi Li\textsuperscript{\maplogo2\thanks{\enspace The authors contributed equally to this work.}}\quad 
Ge Zhang\textsuperscript{\maplogo1,3\samethanks[1]}\quad
Xingwei Qu\textsuperscript{\maplogo2\samethanks[1]}\quad 
Jiali Li\textsuperscript{4}\quad  
Zhaoqun Li\textsuperscript{5}\quad  
Zekun Wang\textsuperscript{\maplogo6}\quad 
}
\\
{\scriptsize \textbf{
Hao Li\textsuperscript{2}\quad  
Ruibin Yuan\textsuperscript{\maplogo7}\quad 
Yinghao Ma\textsuperscript{\maplogo8}\quad 
Kai Zhang\textsuperscript{9}\quad 
Wangchunshu Zhou\textsuperscript{10}\quad
Yiming Liang\textsuperscript{11,12}\quad
}}
\\
{\scriptsize 
\textbf{
Lei Zhang\textsuperscript{1}\ \ 
Lei Ma\textsuperscript{13}\ \ 
Jiajun Zhang\textsuperscript{11,12}\ \ 
Zuowen Li\textsuperscript{14}\ \ 
Stephen W. Huang\textsuperscript{15}\ \ 
Chenghua Lin\textsuperscript{\maplogo2\thanks{\enspace Corresponding authors.}}\ \ 
Jie Fu\textsuperscript{\maplogo7\samethanks[2]}}
}
\\ 
{\tiny \normalfont 
\textsuperscript{1}Stardust.AI\ \ 
\textsuperscript{\maplogo}m-a-p.ai\ \ 
\textsuperscript{2}University of Manchester\ \ 
\textsuperscript{3}University of Waterloo
\textsuperscript{4}National University of Singapore\ \ 
\textsuperscript{5}Zhejiang University\ \
\textsuperscript{6}Beihang University\ \ 
}
\\
{\tiny
\textsuperscript{7}HKUST\ \ 
\textsuperscript{8}Queen Mary University of London\ \ 
\textsuperscript{9}Ohio State University\ \ 
\textsuperscript{10}AIWaves Inc.\ \ 
\textsuperscript{11}Institute of Automation, Chinese Academy of Sciences\ \ 
}
\\
{\tiny
\textsuperscript{12}School of Artificial Intelligence, Chinese Academy of Sciences\ \ 
\textsuperscript{13}Peking University\ \ 
\textsuperscript{14}Beijing Foreign Studies University\ \ 
\textsuperscript{15}harmony.ai\ \ 
}
}

\begin{document}
\maketitle

\begin{abstract}

The advancement of large language models (LLMs) has enhanced the ability to generalize across a wide range of unseen natural language processing (NLP) tasks through instruction-following.
Yet, their effectiveness often diminishes in less-trained languages like Chinese, exacerbated by biased evaluations from data leakage, casting doubt on their true generalizability to new linguistic territories. 
In response, we introduce the Chinese Instruction-Following Benchmark (\textbf{CIF-Bench}), designed to evaluate the zero-shot generalizability of LLMs to the Chinese language. 
CIF-Bench comprises 150 tasks and 15,000 input-output pairs, developed by native speakers to test complex reasoning and Chinese cultural nuances across 20 categories. 
To mitigate data contamination, we release only half of the dataset publicly, with the remainder kept private, and introduce diversified instructions to minimize score variance, totaling 45,000 data instances.
Our evaluation of $28$ selected LLMs reveals a noticeable performance gap, with the best model scoring only 52.9\%, highlighting the limitations of LLMs in less familiar language and task contexts.
This work aims to uncover the current limitations of LLMs in handling Chinese tasks, pushing towards the development of more culturally informed and linguistically diverse models with the released data and benchmark\footnote{\enspace \scriptsize\url{https://yizhilll.github.io/CIF-Bench/}}.

\end{abstract}

\begin{figure}
    \centering
    \includegraphics[width=0.8\linewidth]{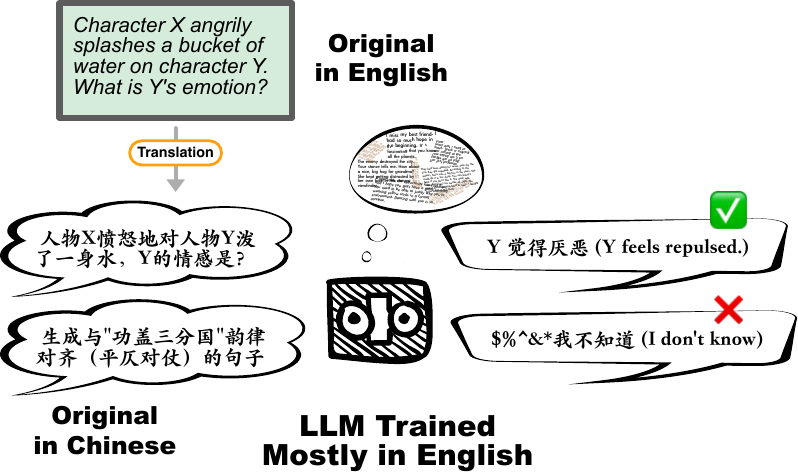}
    \caption{A large language model can tackle English task translated to Chinese, but fail to respond to instruction originally in Chinese.}
    \label{fig:CIF framework}
\end{figure}
\
\section{Introduction}

The landscape of natural language processing (NLP) has been dramatically reshaped by the emergence of large language models (LLMs), which have demonstrated an ability to generalize across unseen NLP tasks, often showcased through the framework of instruction-following~\cite{mishra2021cross,sanh2021T0,wei2021FLAN}. 
Despite these advances, skepticism remains regarding the transferability of this instruction-following capability, particularly in multilingual contexts. The models perform worse when switching to Chinese due to the prevalence of English training data~\cite{huang2023ceval, zhang2023multilingual}, as figured in Fig.~\ref{fig:CIF framework}.
This concern is exacerbated by observations that benchmarks designed to assess the capabilities of LLMs may inadvertently suffer from biased evaluations due to data leakage~\cite{sainz2023nlp_eval_trouble}, particularly when web-scale datasets are employed to enhance model generalizability~\cite{raffel2023exploring}. 
Such observations raise a critical question: While the generalizability of LLMs appears intriguing, do these models face significant challenges when evaluated on private and diversified instruction-formatted tasks in less common language contexts?

To answer this question, we introduce the \textbf{C}hinese \textbf{I}nstruction-\textbf{F}ollowing \textbf{Bench}mark (\textbf{CIF-Bench}), a novel benchmark designed for the zero-shot generalizability evaluation of LLMs, with Chinese serving as an insightful example for multilingual transferred instruction-following tasks. 
Our benchmark comprises $150$ tasks and $15,000$ input-output pairs, with the assistance of native speaker annotators, ensuring the inclusion of human-authored tasks that are not only challenging but also naturally expressed. 
A significant portion ($38.7\%$) of these tasks are designed to test a model's complex natural language inference (NLI) and reasoning capabilities, as well as drawing upon Chinese culture spread across 20 distinct categories. 
In an effort to mitigate future evaluation biases from data leakage, we decide to publicly release only half of the data instances, reserving the rest as a private dataset to maintain an impartial benchmark. 
Furthermore, CIF-Bench enhances its robustness by introducing 5 variations of instructions per task, using these to diminish score variance in private split evaluations as discussed in \S\ref{sec:instruction_diversity}.
CIF-Bench also pioneers a model-based automatic pipeline designed to tackle the inherent challenges of evaluating open-ended natural language generation outputs~\cite{gehrmann2021gem}.

By selecting a range of popular LLMs that support Chinese for evaluation, we aim to depict the limits of current instruction-following capabilities in language transfer contexts as the many models follow an English-oriented pre-training paradigm~\cite{huang2023ceval}.
Our findings reveal that even the best-performing model achieves a score of only $52.9\%$ on CIF-Bench, underscoring the gap that exists when LLMs are confronted with tasks in a less-familiar language and unseen data instances. 
We find that this performance decrement is particularly noticeable in scenarios involving unseen tasks and unseen input-output pairs, contrasting with the models' performance on existing Chinese datasets and translated English-language tasks. 
Such results suggest that while LLMs exhibit impressive generalizability in a context more aligned with observed data, their effectiveness diminishes when faced with the dual challenges of unacquainted languages and novel tasks.

To summarize our contributions, we: 
\begin{itemize}[topsep=2pt,itemsep=-1ex,partopsep=1ex,parsep=1ex]
    \item Present a new benchmark that addresses a critical gap in existing NLP research by focusing on the generalizability of LLMs to an underrepresented language in terms of training and evaluation resources; 
    \item Construct an instruction-following evaluation dataset with $150$ tasks and $45,000$ data samples, and release half of the input-output pairs for future LLM evaluation research; 
    \item Provide an in-depth analysis of 28 LLMs, revealing their limitations in adapting to less familiar languages and task contexts, offering insights into where improvements are needed for instruction-following generalizability.
\end{itemize}

\section{Related Work}

\paragraph{Instruction-Following Evaluation.}

Large-scale pre-trained language models have been found that they can generalize across unseen tasks by fine-tuned on formatted task instructions~\citep{khashabi2020unifiedqa, mishra2021cross, wei2021FLAN, sanh2021T0}.
Early studies attempt to fine-tune and evaluate such a capability in a few-shot manner by providing input-output examples~\cite{ye2021crossfit,  mishra2021cross}.
Following that, another line of research~\citet{bach2022promptsource, wang2022super} ~\citet{bai2024mtbench101} improves the evaluation reliability from the perspective of scaling the task quantity and providing well-defined corresponding instructions.
A more recent concurrent work FollowBench proposes to craft multiple instructions for a single task to evaluate the LLMs, similar to CIF-Bench. A core distinction between CIF-Bench and the FollowBench is that we focus on assessing whether models can stably perform given diversely expressed, but semantically identical instructions, while FollowBench aims to extend the basic instruction with different additional requirements.

\paragraph{Chinese LLM Benchmarks.} 

There have been important efforts, such as CLUE~\citep{xu2020clue} and CUGE~\citep{yao2021cuge}, made to evaluate the pre-trained language on extensive tasks in the Chinese context, which consider the traditional taxonomy of natural language understanding and generation.
As these benchmarks are restricted in the prediction formats and could not fully measure the cross-task generalization of LLMs in the free-form outputs, more recent studies~\cite{huang2023ceval, li2023cmmlu} propose to reformat the tasks into multi-choice question answering, mostly examining the knowledge-base abilities in Chinese.
However, such a strict format could impede the models from fully generalizing to more complex reasoning and creative tasks.
Thereby, we argue that there is a lag in evaluating LLMs instruction-following capacity in the Chinese language.

\section{The Challenging Chinese Instruction-Following Benchmark}\label{sec:benchmark}
The Challenging Chinese Instruction-Following Benchmark unifies the NLP tasks in the prompt-based instruction-following schema~\cite{mishra2021cross} and evaluates the LLMs in a zero-shot manner, which is to say that the models are expected to directly provide the correct \ul{output} given the concatenation of the task \ul{instruction} and data \ul{input} texts. 
Formally, for each data sample in CIF-Bench, the three components we refer to are:
\begin{itemize}[topsep=2pt,itemsep=-1ex,partopsep=1ex,parsep=1ex]
    \item An \ul{instruction} that is provided as the introductory information for a specific NLP task, which is an implicit definition of a ``mapping function" (i.e., task background context) that must be interpreted by the models before proceeding. 
    \item An span of \ul{input} text that encompasses the context to define the specific task scenario. 
    \item A reference as the (potentially) standard \ul{output} in the data instance.
\end{itemize}

\begin{table}[!htp]\centering
\caption{The Statistics of CIF-Bench instruction data. \#Instruction and \#Input-Output refer to the quantity of examples contained in each task.}\label{tab: dataset stats}
\small
\begin{tabular}{lcc}\toprule
\textbf{Split$\rightarrow$} &\textbf{Private} &\textbf{Public} \\\midrule
\#Task &\multicolumn{2}{c}{150} \\
\#Instruction &5 &1 \\
\#Input-Output &50 &50 \\
\midrule
Total Instances &37,500 &7,500 \\
\bottomrule
\end{tabular}
\end{table}

We define a total of 150 curated tasks, constructed according to Chinese linguistic and societal backgrounds, as well as from existing NLP tasks in Chinese and English.
To improve the evaluation robustness, we provide a diversified set of $5$ instructions with the same semantics for each task. 
Considering the potential data leakage issue of LLM benchmarks, we split two halves of $100$ input-output pairs in each task into \textit{private} and \textit{public} partitions, and only test and release the \textit{public} split which contains one instruction variant. 
In sum, there are $45,000$ human-annotated [instruction, input, output] instances produced in CIF-Bench, as suggested in \autoref{tab: dataset stats}. In addition, we provide detailed instructions for all the tasks in Appendix~\ref{apdx:tasks}.

\begin{table}[!htp]\centering
\caption{The statistics of existing and newly designed Tasks. The existing tasks and instances include those translated from English as well as original Chinese data. }\label{tab:new_tasks_stats}
\scriptsize
\begin{tabular}{l|cc}\toprule
\multirow{2}{*}{\textbf{Task}} & \multicolumn{2}{c}{\textbf{Instance}} \\
&\textbf{Existing} & \textbf{Newly Annotated} \\
\midrule
Existing (113) &5,650 & 5,650 \\
Newly Designed (37) &N/A &3,700 \\
\midrule
\textbf{Total} & 5,650 & 9,350 \\
\bottomrule
\end{tabular}
\end{table}

\subsection{Data Collection.}

\paragraph{Collecting Sources.} 
CIF-Bench is designed for the extensive evaluation of Chinese comprehension and generation capabilities in LLMs, particularly focusing on aspects such as creative generation and linguistic abilities that existing benchmarks, such as C-Eval~\cite{huang2023ceval} and C-MMLU~\cite{li2023cmmlu}, struggle to assess. 
First, we select 113 diverse existing English NLP tasks, as shown in \autoref{tab: dataset stats} from Super Natural Instructions (\textbf{SNI})~\cite{wang2022super} and other research work (full list in Appendix~\ref{apdx:tasks}).
We then describe these task instructions in Chinese and a semantically balanced distributed subset from each original English NLP task as the \textit{\textbf{Public}} split of CIF-Bench.
We further ask expert native Chinese speakers, who minimally have undergraduate degrees, to annotate 100 samples per task based on the translated task instructions.
These samples are further deduplicated according to their semantic embeddings.
We finally select 50 samples per task as the \textit{\textbf{Private}} split of CIF-Bench, to guarantee each sample's validity and the balanced label distribution of each task.

\paragraph{Annotation Protocol.} To be specific, we set up a robust three-stage pipeline in our annotation process.
In \textit{\textbf{stage 1}}, to ensure high annotation quality, we hire native speakers with college backgrounds to annotate the data samples in the form of triplet <\ul{instruction}, \ul{input}, \ul{output}> in cooperated with the annotation platform Stardust\footnote{\url{https://stardust.ai}}.
In \textit{\textbf{stage 2}}, the data annotation specialists from the platform conduct a second round of checking on the quality of the samples. The specialists first use the GPT-4 as an auxiliary verification, and the samples scored lower than 6 out of 10 would be directly deleted. The specialists then manually check on the rest of the samples and deleted the unqualified ones. Next, annotators from the \textit{stage 1} would continue the annotation until collecting 100 input-output pairs per task. The specialists also check on the distribution of the labels and answers, to avoid similar input-output pairs for the task.
In \textit{\textbf{stage 3}}, four researchers with NLP backgrounds conduct a final check by inspecting randomly sampled 20 data points from the 150 tasks. If one of the samples does not satisfy the annotation requirements, the task will be returned to the beginning of the annotation pipeline until it passes verification.
Such a pipeline of three stages costs approximately \$24K. 

\paragraph{Detailed Categories.} To further improve CIF-Bench's task diversity, we create 37 additional new tasks and state the related Chinese instructions.
Specifically, we focus on adding Chinese tasks about \textbf{Creative Natural Language Generation}, \textbf{Traditional Chinese}, and \textbf{Complex Role-Playing Text Games}.  
We ask the expert native speakers to annotate 200 samples per task based on the translated task instructions.
These samples are deduplicated and we further select the \textit{\textbf{Public}} and \textit{\textbf{Private}} split from it.
Each task is further annotated with 4 \textit{\textbf{Private}} paraphrased instructions to test whether LLMs understand the Chinese instructions' meanings or overfit to the instructions in the \textit{\textbf{Public}} split.
Each sample and instruction is manually verified or written by the authors to make sure that CIF-Bench is reliable.

\begin{figure}[hbtp]
    \centering
    \includegraphics[width=0.75\linewidth]{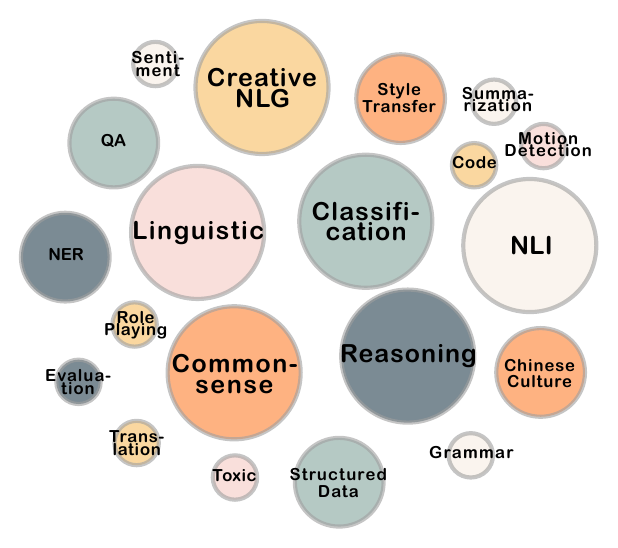}
    \caption{Task Category Distribution in CIF-Bench. The radii have three groups, determined by the number of tasks contained ($\leq10$, $\leq20$, and $>20$).}
    \label{fig: task category stats}
\end{figure}

\subsection{Task Category}
Whilst diverse tasks are provided in CIF-Bench, it would be difficult to analyze the extensive scores from all of the tasks.
By reviewing and summarizing the existing NLP tasks and instruction-following benchmarks, we accordingly categorize the $150$ tasks into \textbf{20} basic types in a multi-label fashion (i.e., a task can be belong to more than one category).
Each category consists of $2$ to $36$ tasks and the quantity distribution is revealed in \autoref{fig: task category stats}. 
Other than the $36$ ``commonsense'' tasks requiring a wide-ranging knowledge base, there are two dominant categories that aim to challenge the logical reasoning abilities of LLMs in CIF-Bench, including $30$ ''natural language inference (NLI)'' and $29$ ``reasoning'' tasks.
In particular, there are 18 tasks designed to require knowledge of unique Chinese cultural contexts.
We describe the definition of each category and the task numbers in Appendix~\ref{apdx:categories}.

\begin{figure}
    \centering
    \includegraphics[width=0.75\linewidth]{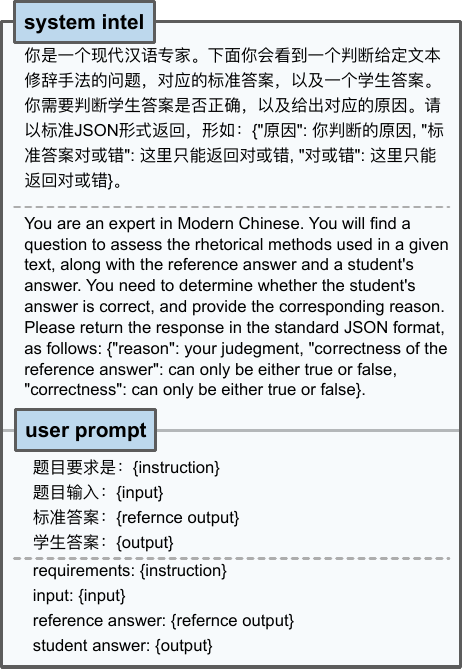}
    \caption{An Exemplar Prompt for GPT-4 Evaluator for the Task ``Chinese Rhetoric Detection''.}
    \label{fig:gpt4evaluator_prompt}                  
\end{figure}

\begin{table*}[!h]\centering
\caption{Overall results in CIF-Bench \textit{Private} split with diversified instructions (1/2). The first column is the average score across \textit{all} the tasks, and the other columns are average scores grouped by task categories. The cells are highlighted with fading colors from \colorbox[HTML]{fbfada}{maximum} to \colorbox[HTML]{b4b4b8}{minimum} in a column.}\label{tab: overall result 1 private}
\scriptsize
\scalebox{0.8}{
\begin{tabular}{lrrrrrrrrrrrr}\toprule
\multicolumn{1}{l}{\textbf{Model Name}} &\multicolumn{1}{p{0.8cm}}{\textbf{Overall}} &\multicolumn{1}{p{1.0cm}}{\textbf{Chinese Culture}} &\multicolumn{1}{p{1.2cm}}{\textbf{Classification}} &\multicolumn{1}{p{0.8cm}}{\textbf{Code}} &\multicolumn{1}{p{1.2cm}}{\textbf{Commonsense}} &\multicolumn{1}{p{1.1cm}}{\textbf{Creative NLG}} &\multicolumn{1}{p{1.0cm}}{\textbf{Evaluation}} &\multicolumn{1}{p{1.0cm}}{\textbf{Grammar}} &\multicolumn{1}{p{1.0cm}}{\textbf{Linguistic}} &\multicolumn{1}{p{1.0cm}}{\textbf{Motion Detection}} &\multicolumn{1}{p{0.8cm}}{\textbf{NER}} \\\midrule
Baichuan2-13B-Chat &\cellcolor[HTML]{fbfada}0.529 &\cellcolor[HTML]{fbfada}0.520 &\cellcolor[HTML]{fbfada}0.674 &\cellcolor[HTML]{eae8d9}0.333 &\cellcolor[HTML]{fbfada}0.641 &\cellcolor[HTML]{f7f6d9}0.497 &\cellcolor[HTML]{fbfada}0.686 &\cellcolor[HTML]{fbfada}0.542 &\cellcolor[HTML]{fbfada}0.528 &\cellcolor[HTML]{fbfada}0.578 &\cellcolor[HTML]{fbfada}0.563 \\
Qwen-72B-Chat &\cellcolor[HTML]{f9f8d9}0.519 &\cellcolor[HTML]{f7f6d9}0.486 &\cellcolor[HTML]{f6f5d9}0.630 &\cellcolor[HTML]{e0ded7}0.296 &\cellcolor[HTML]{faf9d9}0.634 &\cellcolor[HTML]{fbfada}0.508 &\cellcolor[HTML]{efeed9}0.634 &\cellcolor[HTML]{f2f1d9}0.458 &\cellcolor[HTML]{f9f8d9}0.520 &\cellcolor[HTML]{efedd9}0.494 &\cellcolor[HTML]{faf8d9}0.550 \\
Yi-34B-Chat &\cellcolor[HTML]{f8f7d9}0.512 &\cellcolor[HTML]{f7f6d9}0.483 &\cellcolor[HTML]{f4f3d9}0.606 &\cellcolor[HTML]{efeed9}0.347 &\cellcolor[HTML]{f9f8d9}0.623 &\cellcolor[HTML]{f7f6d9}0.497 &\cellcolor[HTML]{e7e5d9}0.598 &\cellcolor[HTML]{f4f3d9}0.480 &\cellcolor[HTML]{f5f4d9}0.490 &\cellcolor[HTML]{faf9d9}0.575 &\cellcolor[HTML]{f8f7d9}0.525 \\
Qwen-14B-Chat &\cellcolor[HTML]{f7f6d9}0.500 &\cellcolor[HTML]{f7f6d9}0.481 &\cellcolor[HTML]{f2f0d9}0.582 &\cellcolor[HTML]{e2e0d8}0.307 &\cellcolor[HTML]{f8f7d9}0.614 &\cellcolor[HTML]{f6f5d9}0.494 &\cellcolor[HTML]{f1f0d9}0.645 &\cellcolor[HTML]{efeed9}0.428 &\cellcolor[HTML]{f3f1d9}0.475 &\cellcolor[HTML]{efeed9}0.496 &\cellcolor[HTML]{f7f6d9}0.513 \\
Deepseek-LLM-67B-Chat &\cellcolor[HTML]{f3f2d9}0.471 &\cellcolor[HTML]{f6f5d9}0.467 &\cellcolor[HTML]{f1efd9}0.571 &\cellcolor[HTML]{dad8d2}0.259 &\cellcolor[HTML]{f4f3d9}0.577 &\cellcolor[HTML]{f4f3d9}0.486 &\cellcolor[HTML]{dfddd6}0.549 &\cellcolor[HTML]{f1efd9}0.442 &\cellcolor[HTML]{f3f2d9}0.476 &\cellcolor[HTML]{ecebd9}0.475 &\cellcolor[HTML]{f6f5d9}0.509 \\
Baichuan-13B-Chat &\cellcolor[HTML]{f1efd9}0.450 &\cellcolor[HTML]{f1efd9}0.408 &\cellcolor[HTML]{e9e7d9}0.491 &\cellcolor[HTML]{dedcd5}0.286 &\cellcolor[HTML]{f2f1d9}0.552 &\cellcolor[HTML]{e6e4d9}0.439 &\cellcolor[HTML]{f7f6d9}0.670 &\cellcolor[HTML]{eeedd9}0.417 &\cellcolor[HTML]{ebe9d9}0.422 &\cellcolor[HTML]{edecd9}0.482 &\cellcolor[HTML]{f5f4d9}0.486 \\
Chatglm3-6B &\cellcolor[HTML]{efedd9}0.436 &\cellcolor[HTML]{eeedd9}0.381 &\cellcolor[HTML]{e4e2d9}0.439 &\cellcolor[HTML]{e9e7d9}0.330 &\cellcolor[HTML]{f1f0d9}0.541 &\cellcolor[HTML]{eae8d9}0.452 &\cellcolor[HTML]{e3e1d9}0.577 &\cellcolor[HTML]{e4e2d9}0.310 &\cellcolor[HTML]{e0ded7}0.358 &\cellcolor[HTML]{e7e5d9}0.436 &\cellcolor[HTML]{f2f1d9}0.453 \\
Yi-6B-Chat &\cellcolor[HTML]{edebd9}0.417 &\cellcolor[HTML]{f0efd9}0.402 &\cellcolor[HTML]{e5e4d9}0.454 &\cellcolor[HTML]{e3e1d9}0.313 &\cellcolor[HTML]{efeed9}0.523 &\cellcolor[HTML]{e1dfd8}0.425 &\cellcolor[HTML]{d8d7d1}0.506 &\cellcolor[HTML]{ebe9d9}0.383 &\cellcolor[HTML]{e5e3d9}0.383 &\cellcolor[HTML]{eeecd9}0.487 &\cellcolor[HTML]{eeedd9}0.396 \\
Baichuan2-7B-Chat &\cellcolor[HTML]{ecead9}0.412 &\cellcolor[HTML]{f3f2d9}0.437 &\cellcolor[HTML]{f8f7d9}0.647 &\cellcolor[HTML]{cac9c7}0.160 &\cellcolor[HTML]{efeed9}0.520 &\cellcolor[HTML]{d8d7d1}0.402 &\cellcolor[HTML]{e3e1d9}0.580 &\cellcolor[HTML]{f7f6d9}0.511 &\cellcolor[HTML]{eeedd9}0.444 &\cellcolor[HTML]{e9e8d9}0.455 &\cellcolor[HTML]{efedd9}0.407 \\
Chatglm2-6B &\cellcolor[HTML]{e4e2d9}0.352 &\cellcolor[HTML]{e5e3d9}0.278 &\cellcolor[HTML]{e7e5d9}0.469 &\cellcolor[HTML]{efedd9}0.346 &\cellcolor[HTML]{e4e2d9}0.403 &\cellcolor[HTML]{e1dfd7}0.424 &\cellcolor[HTML]{dcdbd4}0.535 &\cellcolor[HTML]{dbdad3}0.274 &\cellcolor[HTML]{e7e6d9}0.397 &\cellcolor[HTML]{e3e1d9}0.406 &\cellcolor[HTML]{e2e0d8}0.240 \\
Chatglm-6B-Sft &\cellcolor[HTML]{e4e2d9}0.349 &\cellcolor[HTML]{e4e2d9}0.265 &\cellcolor[HTML]{e5e4d9}0.454 &\cellcolor[HTML]{f6f4d9}0.365 &\cellcolor[HTML]{e2e0d8}0.385 &\cellcolor[HTML]{edebd9}0.462 &\cellcolor[HTML]{dfddd6}0.554 &\cellcolor[HTML]{e2e0d8}0.296 &\cellcolor[HTML]{e5e3d9}0.379 &\cellcolor[HTML]{e6e4d9}0.427 &\cellcolor[HTML]{dedcd5}0.232 \\
Chinese-Llama2-Linly-13B &\cellcolor[HTML]{e3e1d9}0.344 &\cellcolor[HTML]{e3e1d9}0.250 &\cellcolor[HTML]{e6e4d9}0.462 &\cellcolor[HTML]{e2e0d8}0.311 &\cellcolor[HTML]{e4e2d9}0.399 &\cellcolor[HTML]{e3e1d9}0.429 &\cellcolor[HTML]{e0ded7}0.557 &\cellcolor[HTML]{dbd9d3}0.273 &\cellcolor[HTML]{e0ded6}0.358 &\cellcolor[HTML]{dddcd5}0.385 &\cellcolor[HTML]{e4e3d9}0.268 \\
GPT-3.5-Turbo-Sft &\cellcolor[HTML]{e3e1d9}0.343 &\cellcolor[HTML]{e4e2d9}0.269 &\cellcolor[HTML]{e3e1d9}0.427 &\cellcolor[HTML]{e0ded7}0.298 &\cellcolor[HTML]{e3e1d9}0.389 &\cellcolor[HTML]{d5d4cf}0.395 &\cellcolor[HTML]{e2e0d8}0.575 &\cellcolor[HTML]{e5e3d9}0.325 &\cellcolor[HTML]{e3e1d9}0.365 &\cellcolor[HTML]{deddd6}0.389 &\cellcolor[HTML]{dbd9d3}0.226 \\
Chinese-Alpaca-2-13B &\cellcolor[HTML]{e3e1d9}0.341 &\cellcolor[HTML]{dedcd6}0.242 &\cellcolor[HTML]{e2e0d8}0.421 &\cellcolor[HTML]{f3f1d9}0.356 &\cellcolor[HTML]{e0ded7}0.382 &\cellcolor[HTML]{e7e5d9}0.442 &\cellcolor[HTML]{e8e6d9}0.602 &\cellcolor[HTML]{d6d5d0}0.256 &\cellcolor[HTML]{e2e0d8}0.363 &\cellcolor[HTML]{e6e4d9}0.430 &\cellcolor[HTML]{d3d2ce}0.210 \\
Chinese-Alpaca-13B &\cellcolor[HTML]{e1dfd7}0.334 &\cellcolor[HTML]{e2e0d8}0.250 &\cellcolor[HTML]{dddbd4}0.399 &\cellcolor[HTML]{f0eed9}0.348 &\cellcolor[HTML]{d9d7d2}0.364 &\cellcolor[HTML]{e5e3d9}0.435 &\cellcolor[HTML]{ebe9d9}0.616 &\cellcolor[HTML]{dcdad4}0.275 &\cellcolor[HTML]{dbdad4}0.349 &\cellcolor[HTML]{e5e3d9}0.421 &\cellcolor[HTML]{d9d8d2}0.223 \\
Chinese-Alpaca-7B &\cellcolor[HTML]{e0dfd7}0.334 &\cellcolor[HTML]{d1cfcc}0.216 &\cellcolor[HTML]{e0ded6}0.412 &\cellcolor[HTML]{fbfada}0.378 &\cellcolor[HTML]{e0ded7}0.381 &\cellcolor[HTML]{e1dfd8}0.425 &\cellcolor[HTML]{e2e0d8}0.576 &\cellcolor[HTML]{d9d7d2}0.265 &\cellcolor[HTML]{e0ded7}0.359 &\cellcolor[HTML]{dfded6}0.393 &\cellcolor[HTML]{e3e1d9}0.243 \\
Chinese-Llama2-Linly-7B &\cellcolor[HTML]{e0ded7}0.333 &\cellcolor[HTML]{d2d0cd}0.218 &\cellcolor[HTML]{e5e3d9}0.451 &\cellcolor[HTML]{e9e7d9}0.330 &\cellcolor[HTML]{e3e1d9}0.396 &\cellcolor[HTML]{e2e0d8}0.427 &\cellcolor[HTML]{e4e2d9}0.583 &\cellcolor[HTML]{d4d2ce}0.248 &\cellcolor[HTML]{dcdad4}0.350 &\cellcolor[HTML]{e3e1d9}0.410 &\cellcolor[HTML]{dddcd5}0.231 \\
Tigerbot-13B-Chat &\cellcolor[HTML]{dfddd6}0.331 &\cellcolor[HTML]{cbcac8}0.205 &\cellcolor[HTML]{dcdad4}0.397 &\cellcolor[HTML]{e2e0d8}0.309 &\cellcolor[HTML]{e2e0d8}0.385 &\cellcolor[HTML]{dfddd6}0.420 &\cellcolor[HTML]{ebe9d9}0.614 &\cellcolor[HTML]{e4e2d9}0.310 &\cellcolor[HTML]{e5e3d9}0.379 &\cellcolor[HTML]{d2d1cd}0.341 &\cellcolor[HTML]{e5e3d9}0.276 \\
Telechat-7B &\cellcolor[HTML]{dedcd6}0.329 &\cellcolor[HTML]{e4e2d9}0.267 &\cellcolor[HTML]{cdccca}0.338 &\cellcolor[HTML]{e6e4d9}0.321 &\cellcolor[HTML]{e6e4d9}0.420 &\cellcolor[HTML]{d9d7d2}0.404 &\cellcolor[HTML]{cccbc8}0.420 &\cellcolor[HTML]{dbd9d3}0.272 &\cellcolor[HTML]{b4b4b8}0.265 &\cellcolor[HTML]{cecdca}0.327 &\cellcolor[HTML]{e8e7d9}0.320 \\
Ziya-Llama-13B &\cellcolor[HTML]{dedcd5}0.329 &\cellcolor[HTML]{c6c5c4}0.196 &\cellcolor[HTML]{dddbd5}0.402 &\cellcolor[HTML]{e7e5d9}0.324 &\cellcolor[HTML]{cfcecb}0.341 &\cellcolor[HTML]{e3e1d9}0.428 &\cellcolor[HTML]{ebead9}0.616 &\cellcolor[HTML]{e4e2d9}0.312 &\cellcolor[HTML]{dcdad4}0.349 &\cellcolor[HTML]{e1dfd8}0.400 &\cellcolor[HTML]{dcdad4}0.228 \\
Chinese-Alpaca-33B &\cellcolor[HTML]{dddbd4}0.326 &\cellcolor[HTML]{dad8d3}0.234 &\cellcolor[HTML]{d5d4cf}0.370 &\cellcolor[HTML]{f8f7d9}0.372 &\cellcolor[HTML]{d9d7d2}0.364 &\cellcolor[HTML]{e3e1d9}0.429 &\cellcolor[HTML]{ebe9d9}0.614 &\cellcolor[HTML]{d3d2ce}0.246 &\cellcolor[HTML]{cdccc9}0.318 &\cellcolor[HTML]{dbdad3}0.377 &\cellcolor[HTML]{d8d7d1}0.221 \\
Tigerbot-7B-Chat &\cellcolor[HTML]{dcdad4}0.325 &\cellcolor[HTML]{d1d0cc}0.218 &\cellcolor[HTML]{dbdad4}0.395 &\cellcolor[HTML]{e1e0d8}0.306 &\cellcolor[HTML]{dbdad3}0.370 &\cellcolor[HTML]{dcdbd4}0.413 &\cellcolor[HTML]{efedd9}0.631 &\cellcolor[HTML]{e1dfd8}0.294 &\cellcolor[HTML]{e3e1d9}0.370 &\cellcolor[HTML]{d9d7d2}0.368 &\cellcolor[HTML]{d5d4cf}0.215 \\
Chinese-Alpaca-2-7B &\cellcolor[HTML]{dbd9d3}0.323 &\cellcolor[HTML]{d0cfcc}0.215 &\cellcolor[HTML]{d6d5d0}0.374 &\cellcolor[HTML]{ebe9d9}0.335 &\cellcolor[HTML]{d9d8d2}0.366 &\cellcolor[HTML]{dddcd5}0.415 &\cellcolor[HTML]{dedcd5}0.546 &\cellcolor[HTML]{d6d5d0}0.257 &\cellcolor[HTML]{d1cfcc}0.326 &\cellcolor[HTML]{e0ded7}0.395 &\cellcolor[HTML]{d5d4cf}0.215 \\
Aquilachat-7B &\cellcolor[HTML]{d4d3ce}0.309 &\cellcolor[HTML]{b4b4b8}0.162 &\cellcolor[HTML]{b4b4b8}0.234 &\cellcolor[HTML]{dfddd6}0.291 &\cellcolor[HTML]{c6c5c4}0.320 &\cellcolor[HTML]{e5e3d9}0.437 &\cellcolor[HTML]{c1c0c1}0.344 &\cellcolor[HTML]{b4b4b8}0.135 &\cellcolor[HTML]{b4b4b8}0.266 &\cellcolor[HTML]{cac9c7}0.309 &\cellcolor[HTML]{e6e4d9}0.287 \\
Moss-Moon-003-Sft &\cellcolor[HTML]{d1cfcc}0.302 &\cellcolor[HTML]{cfcecb}0.214 &\cellcolor[HTML]{dedcd5}0.405 &\cellcolor[HTML]{dcdad4}0.274 &\cellcolor[HTML]{d1d0cc}0.347 &\cellcolor[HTML]{cfcecb}0.380 &\cellcolor[HTML]{d0cfcb}0.448 &\cellcolor[HTML]{e3e1d9}0.305 &\cellcolor[HTML]{d8d6d1}0.341 &\cellcolor[HTML]{dcdad4}0.378 &\cellcolor[HTML]{dedcd5}0.232 \\
Qwen-7B-Chat &\cellcolor[HTML]{d0cfcc}0.301 &\cellcolor[HTML]{cdccca}0.211 &\cellcolor[HTML]{dfddd6}0.410 &\cellcolor[HTML]{dfddd6}0.289 &\cellcolor[HTML]{d2d1cd}0.349 &\cellcolor[HTML]{d4d2ce}0.391 &\cellcolor[HTML]{dcdad4}0.531 &\cellcolor[HTML]{cbcac8}0.219 &\cellcolor[HTML]{e6e4d9}0.387 &\cellcolor[HTML]{e2e0d8}0.404 &\cellcolor[HTML]{d2d0cd}0.208 \\
Belle-13B-Sft &\cellcolor[HTML]{bebdbf}0.264 &\cellcolor[HTML]{c7c6c5}0.198 &\cellcolor[HTML]{c6c5c4}0.307 &\cellcolor[HTML]{dedcd5}0.285 &\cellcolor[HTML]{c4c3c3}0.316 &\cellcolor[HTML]{c3c2c2}0.349 &\cellcolor[HTML]{cac9c7}0.409 &\cellcolor[HTML]{d1cfcc}0.237 &\cellcolor[HTML]{c7c6c5}0.305 &\cellcolor[HTML]{b4b4b8}0.222 &\cellcolor[HTML]{c2c1c2}0.177 \\
CPM-Bee-10B &\cellcolor[HTML]{b4b4b8}0.244 &\cellcolor[HTML]{dad8d2}0.234 &\cellcolor[HTML]{d7d6d0}0.377 &\cellcolor[HTML]{b4b4b8}0.024 &\cellcolor[HTML]{b4b4b8}0.278 &\cellcolor[HTML]{b4b4b8}0.311 &\cellcolor[HTML]{b4b4b8}0.255 &\cellcolor[HTML]{e3e1d9}0.302 &\cellcolor[HTML]{bab9bc}0.278 &\cellcolor[HTML]{cfcdcb}0.327 &\cellcolor[HTML]{b4b4b8}0.148 \\
\bottomrule
\end{tabular}
}
\end{table*}

\subsection{Task-based Automatic Evaluation}

As the CIF-Bench aims to provide a comprehensive evaluation of the LLM instruction-following capability, we argue that the metrics should be designed case by case in task granularity to evaluate the open-ended textual outputs, rather than simply reformatting all tasks into choice questions and using the conditional probability to approximate the models' predictions.

After a thorough review of the task instructions, we categorize the output requirements into the four following types and design corresponding task-level metrics.
\textbf{Multi-class Classification}: We use \textbf{accuracy} as the metric if the task requires the model to predict one label from 2 or more classes in the output. 
\textbf{Multi-label Classification}: We use \textbf{F1 score} as the metric if the task requires the model to predict one label from 2 or more classes in the output. 
\textbf{Creative Generation}: Regarding the tasks that have no absolute criteria of the standard answer, we require a model-based evaluator to provide information regarding a given output, including \textbf{creativity}, \textbf{fluency}, the level of \textbf{instruction-following}, and the \textbf{confidence} of the evaluator.
\textbf{Semantic Similarity}: For the remaining tasks that can be evaluated by the semantic similarity between the golden reference and model output, we use a pre-trained language 
All scores used in CIF-Bench either naturally range from $0$ to $1$, or are normalized to the same range.

One core dilemma in evaluating the open-ended instruction-following capabilities of LLMs is that model predictions are hard to verify even with reference answers. For instance, it is intractable to handcraft regex rules to extract the predictions from LLMs for the extensive number of tasks, since the answers could be expressed in various formats, or drowned in redundant contexts like reasoning progress.
Inspired by G-Eval~\cite{liu2023gpteval}, we leverage OpenAI's GPT-4\footnote{https://openai.com/gpt-4} as a relatively reliable evaluator for multi-class classification, multi-label classification, and creative generation tasks, to overcome such issues. The GPT-4 evaluator is prompted to assess the outputs according to the given task instruction and the input-output reference, as shown by the example in ~\autoref{fig:gpt4evaluator_prompt} and the full list of evaluation prompts in Appendix~\ref{apdx:tasks}.
the remaining tasks that can be evaluated by the semantic similarity between the golden reference and model output, we use a lightweight multilingual encoder, BLEURT~\cite{sellam2020bleurt}, to measure the relevance between the reference and LLM output.

Given a set of task instructions $I$, we denote the performance score of model $m$ on task $t$ as:
$$
S^{m}_{t}=\frac{1}{|D_t|}\sum_{d \in D_t}\frac{1}{|I|}\sum_{i \in I}{{s^{m}_{t}(i,d)}}
$$
, where $D_t$ refers to the set of data samples for task $t$.
In the case of the \textit{public} split, the instruction set $I$ is reduced to one single element. 
In we take the average of task-level scores $\overline{S^m}$ as the indicator of overall performance for a model $m$.

\begin{table*}[!hbtp]\centering
\caption{Overall results in CIF-Bench \textit{Private} split with diversified Instructions (2/2). The first column is the average score across \textit{all} the tasks, and the rest columns are average scores grouped by task categories. The cells are highlighted with fading colors from \colorbox[HTML]{fbfada}{maximum} to \colorbox[HTML]{b4b4b8}{minimum} in a column.}\label{tab: overall result 2 private}
\scriptsize
\scalebox{0.8}{
\begin{tabular}{lrrrrrrrrrrrr}\toprule
\textbf{Model Name} &\textbf{Overall} &\multicolumn{1}{p{0.6cm}}{\textbf{NLI}} &\multicolumn{1}{p{0.6cm}}{\textbf{QA}} &\multicolumn{1}{p{1.0cm}}{\textbf{Reasoning}} &\multicolumn{1}{p{1.0cm}}{\textbf{Role Playing}} &\multicolumn{1}{p{1.0cm}}{\textbf{Sentiment}} &\multicolumn{1}{p{1.0cm}}{\textbf{Structured Data}} &\multicolumn{1}{p{1.0cm}}{\textbf{Style Transfer}} &\multicolumn{1}{p{1.4cm}}{\textbf{Summarization}} & \multicolumn{1}{p{0.8cm}}{\textbf{Toxic}} &\multicolumn{1}{p{1.2cm}}{\textbf{Translation}} \\\midrule
Baichuan2-13B-Chat &\cellcolor[HTML]{fbfada}0.529 &\cellcolor[HTML]{fbfada}0.632 &\cellcolor[HTML]{fbfada}0.569 &\cellcolor[HTML]{f9f8d9}0.515 &\cellcolor[HTML]{f5f4d9}0.752 &\cellcolor[HTML]{fbfada}0.624 &\cellcolor[HTML]{f6f5d9}0.459 &\cellcolor[HTML]{fbfada}0.462 &\cellcolor[HTML]{e6e4d9}0.332 &\cellcolor[HTML]{e3e1d9}0.441 &\cellcolor[HTML]{f4f2d9}0.273 \\
Qwen-72B-Chat &\cellcolor[HTML]{f9f8d9}0.519 &\cellcolor[HTML]{faf9d9}0.626 &\cellcolor[HTML]{faf9d9}0.565 &\cellcolor[HTML]{fbfada}0.528 &\cellcolor[HTML]{faf9d9}0.762 &\cellcolor[HTML]{f9f8d9}0.613 &\cellcolor[HTML]{fbfada}0.496 &\cellcolor[HTML]{faf9d9}0.459 &\cellcolor[HTML]{dfddd6}0.282 &\cellcolor[HTML]{fbfada}0.608 &\cellcolor[HTML]{f3f2d9}0.271 \\
Yi-34B-Chat &\cellcolor[HTML]{f8f7d9}0.512 &\cellcolor[HTML]{f9f8d9}0.619 &\cellcolor[HTML]{f9f8d9}0.554 &\cellcolor[HTML]{f7f6d9}0.494 &\cellcolor[HTML]{f7f6d9}0.757 &\cellcolor[HTML]{f5f4d9}0.580 &\cellcolor[HTML]{f8f6d9}0.472 &\cellcolor[HTML]{f7f6d9}0.439 &\cellcolor[HTML]{e8e6d9}0.346 &\cellcolor[HTML]{edecd9}0.514 &\cellcolor[HTML]{f1efd9}0.259 \\
Qwen-14B-Chat &\cellcolor[HTML]{f7f6d9}0.500 &\cellcolor[HTML]{f9f8d9}0.616 &\cellcolor[HTML]{f8f7d9}0.548 &\cellcolor[HTML]{f8f7d9}0.507 &\cellcolor[HTML]{fbfada}0.764 &\cellcolor[HTML]{f5f4d9}0.583 &\cellcolor[HTML]{f7f6d9}0.469 &\cellcolor[HTML]{f9f8d9}0.453 &\cellcolor[HTML]{dfddd6}0.283 &\cellcolor[HTML]{f6f5d9}0.575 &\cellcolor[HTML]{f1f0d9}0.262 \\
Deepseek-LLM-67B-Chat &\cellcolor[HTML]{f3f2d9}0.471 &\cellcolor[HTML]{f4f3d9}0.566 &\cellcolor[HTML]{f3f2d9}0.496 &\cellcolor[HTML]{f1efd9}0.439 &\cellcolor[HTML]{e3e1d9}0.711 &\cellcolor[HTML]{f1efd9}0.546 &\cellcolor[HTML]{f0efd9}0.409 &\cellcolor[HTML]{f6f5d9}0.436 &\cellcolor[HTML]{dad8d3}0.262 &\cellcolor[HTML]{f5f4d9}0.570 &\cellcolor[HTML]{ecead9}0.235 \\
Baichuan-13B-Chat &\cellcolor[HTML]{f1efd9}0.450 &\cellcolor[HTML]{f4f3d9}0.565 &\cellcolor[HTML]{f4f3d9}0.505 &\cellcolor[HTML]{eae8d9}0.377 &\cellcolor[HTML]{e0ded7}0.704 &\cellcolor[HTML]{f1f0d9}0.552 &\cellcolor[HTML]{edecd9}0.387 &\cellcolor[HTML]{f1efd9}0.402 &\cellcolor[HTML]{e8e6d9}0.350 &\cellcolor[HTML]{e1dfd7}0.431 &\cellcolor[HTML]{faf9d9}0.304 \\
Chatglm3-6B &\cellcolor[HTML]{efedd9}0.436 &\cellcolor[HTML]{f2f1d9}0.544 &\cellcolor[HTML]{f4f2d9}0.503 &\cellcolor[HTML]{eeecd9}0.414 &\cellcolor[HTML]{faf9d9}0.762 &\cellcolor[HTML]{f2f1d9}0.560 &\cellcolor[HTML]{f4f3d9}0.446 &\cellcolor[HTML]{f1efd9}0.402 &\cellcolor[HTML]{e5e3d9}0.321 &\cellcolor[HTML]{d4d2ce}0.391 &\cellcolor[HTML]{f3f2d9}0.270 \\
Yi-6B-Chat &\cellcolor[HTML]{edebd9}0.417 &\cellcolor[HTML]{f0efd9}0.523 &\cellcolor[HTML]{efedd9}0.457 &\cellcolor[HTML]{e9e7d9}0.369 &\cellcolor[HTML]{f6f5d9}0.754 &\cellcolor[HTML]{e9e7d9}0.482 &\cellcolor[HTML]{efedd9}0.401 &\cellcolor[HTML]{edecd9}0.380 &\cellcolor[HTML]{e4e2d9}0.310 &\cellcolor[HTML]{e5e3d9}0.455 &\cellcolor[HTML]{eae8d9}0.227 \\
Baichuan2-7B-Chat &\cellcolor[HTML]{ecead9}0.412 &\cellcolor[HTML]{edebd9}0.489 &\cellcolor[HTML]{e8e7d9}0.395 &\cellcolor[HTML]{edebd9}0.406 &\cellcolor[HTML]{d1cfcc}0.670 &\cellcolor[HTML]{edebd9}0.517 &\cellcolor[HTML]{e8e6d9}0.342 &\cellcolor[HTML]{dad8d2}0.298 &\cellcolor[HTML]{b4b4b8}0.101 &\cellcolor[HTML]{e6e5d9}0.463 &\cellcolor[HTML]{bdbcbe}0.138 \\
Chatglm2-6B &\cellcolor[HTML]{e4e2d9}0.352 &\cellcolor[HTML]{e4e2d9}0.397 &\cellcolor[HTML]{e4e2d9}0.352 &\cellcolor[HTML]{e4e2d9}0.326 &\cellcolor[HTML]{e4e2d9}0.714 &\cellcolor[HTML]{e3e1d9}0.438 &\cellcolor[HTML]{e2e0d8}0.298 &\cellcolor[HTML]{e1dfd7}0.313 &\cellcolor[HTML]{e5e3d9}0.320 &\cellcolor[HTML]{e6e4d9}0.461 &\cellcolor[HTML]{e3e1d9}0.190 \\
Chatglm-6B-Sft &\cellcolor[HTML]{e4e2d9}0.349 &\cellcolor[HTML]{e2e0d8}0.380 &\cellcolor[HTML]{dbdad3}0.321 &\cellcolor[HTML]{dcdbd4}0.292 &\cellcolor[HTML]{e6e4d9}0.718 &\cellcolor[HTML]{dedcd5}0.415 &\cellcolor[HTML]{e2e0d8}0.296 &\cellcolor[HTML]{e5e4d9}0.333 &\cellcolor[HTML]{e8e6d9}0.351 &\cellcolor[HTML]{e3e1d9}0.441 &\cellcolor[HTML]{e3e1d9}0.190 \\
Chinese-Llama2-Linly-13B &\cellcolor[HTML]{e3e1d9}0.344 &\cellcolor[HTML]{e3e1d9}0.390 &\cellcolor[HTML]{dfddd6}0.330 &\cellcolor[HTML]{e2e0d8}0.313 &\cellcolor[HTML]{cac9c7}0.653 &\cellcolor[HTML]{e3e1d9}0.433 &\cellcolor[HTML]{dedcd5}0.279 &\cellcolor[HTML]{e5e3d9}0.332 &\cellcolor[HTML]{e1dfd8}0.292 &\cellcolor[HTML]{e5e4d9}0.457 &\cellcolor[HTML]{dcdad4}0.181 \\
GPT-3.5-Turbo-Sft &\cellcolor[HTML]{e3e1d9}0.343 &\cellcolor[HTML]{e3e1d9}0.382 &\cellcolor[HTML]{e8e7d9}0.394 &\cellcolor[HTML]{e6e4d9}0.345 &\cellcolor[HTML]{e2e0d8}0.710 &\cellcolor[HTML]{e2e0d8}0.433 &\cellcolor[HTML]{e6e4d9}0.324 &\cellcolor[HTML]{c9c9c7}0.266 &\cellcolor[HTML]{e1dfd8}0.290 &\cellcolor[HTML]{d6d4cf}0.397 &\cellcolor[HTML]{eae8d9}0.225 \\
Chinese-Alpaca-2-13B &\cellcolor[HTML]{e3e1d9}0.341 &\cellcolor[HTML]{e0ded7}0.376 &\cellcolor[HTML]{e1dfd7}0.334 &\cellcolor[HTML]{e3e1d9}0.317 &\cellcolor[HTML]{e4e2d9}0.714 &\cellcolor[HTML]{e6e4d9}0.459 &\cellcolor[HTML]{e3e1d9}0.299 &\cellcolor[HTML]{e3e1d9}0.316 &\cellcolor[HTML]{e4e2d9}0.308 &\cellcolor[HTML]{e5e3d9}0.452 &\cellcolor[HTML]{e5e3d9}0.200 \\
Chinese-Alpaca-13B &\cellcolor[HTML]{e1dfd7}0.334 &\cellcolor[HTML]{dddbd5}0.370 &\cellcolor[HTML]{d7d5d0}0.309 &\cellcolor[HTML]{e3e1d9}0.319 &\cellcolor[HTML]{e8e7d9}0.724 &\cellcolor[HTML]{e1dfd7}0.426 &\cellcolor[HTML]{dfddd6}0.285 &\cellcolor[HTML]{deddd6}0.307 &\cellcolor[HTML]{e3e1d9}0.298 &\cellcolor[HTML]{e4e2d9}0.445 &\cellcolor[HTML]{dcdad4}0.181 \\
Chinese-Alpaca-7B &\cellcolor[HTML]{e0dfd7}0.334 &\cellcolor[HTML]{e3e1d9}0.383 &\cellcolor[HTML]{dddcd5}0.326 &\cellcolor[HTML]{dddbd5}0.295 &\cellcolor[HTML]{e2e0d8}0.710 &\cellcolor[HTML]{dcdad4}0.409 &\cellcolor[HTML]{e3e1d9}0.301 &\cellcolor[HTML]{e4e2d9}0.327 &\cellcolor[HTML]{e5e4d9}0.325 &\cellcolor[HTML]{d8d7d1}0.405 &\cellcolor[HTML]{dfddd6}0.186 \\
Chinese-Llama2-Linly-7B &\cellcolor[HTML]{e0ded7}0.333 &\cellcolor[HTML]{dcdad4}0.367 &\cellcolor[HTML]{e3e1d9}0.345 &\cellcolor[HTML]{d8d6d1}0.276 &\cellcolor[HTML]{dddbd5}0.698 &\cellcolor[HTML]{e3e1d9}0.433 &\cellcolor[HTML]{d8d7d1}0.259 &\cellcolor[HTML]{e2e0d8}0.315 &\cellcolor[HTML]{e4e2d9}0.310 &\cellcolor[HTML]{e7e5d9}0.469 &\cellcolor[HTML]{d2d1cd}0.168 \\
Tigerbot-13B-Chat &\cellcolor[HTML]{dfddd6}0.331 &\cellcolor[HTML]{d9d8d2}0.363 &\cellcolor[HTML]{dfddd6}0.329 &\cellcolor[HTML]{dfddd6}0.301 &\cellcolor[HTML]{dbdad3}0.694 &\cellcolor[HTML]{dfddd6}0.419 &\cellcolor[HTML]{dedcd5}0.280 &\cellcolor[HTML]{e0ded7}0.310 &\cellcolor[HTML]{dfddd6}0.283 &\cellcolor[HTML]{d4d3cf}0.393 &\cellcolor[HTML]{dfded6}0.186 \\
Telechat-7B &\cellcolor[HTML]{dedcd6}0.329 &\cellcolor[HTML]{e3e1d9}0.388 &\cellcolor[HTML]{e4e2d9}0.355 &\cellcolor[HTML]{cfcecb}0.244 &\cellcolor[HTML]{d2d0cd}0.672 &\cellcolor[HTML]{cbcac8}0.344 &\cellcolor[HTML]{e7e5d9}0.334 &\cellcolor[HTML]{e6e4d9}0.335 &\cellcolor[HTML]{e3e1d9}0.299 &\cellcolor[HTML]{cac9c8}0.364 &\cellcolor[HTML]{dedcd5}0.184 \\
Ziya-Llama-13B &\cellcolor[HTML]{dedcd5}0.329 &\cellcolor[HTML]{d4d2ce}0.351 &\cellcolor[HTML]{cac9c7}0.279 &\cellcolor[HTML]{e2e0d8}0.313 &\cellcolor[HTML]{e7e5d9}0.721 &\cellcolor[HTML]{e7e5d9}0.468 &\cellcolor[HTML]{e4e2d9}0.311 &\cellcolor[HTML]{d6d5d0}0.291 &\cellcolor[HTML]{dedcd5}0.278 &\cellcolor[HTML]{e1dfd8}0.431 &\cellcolor[HTML]{d7d6d1}0.175 \\
Chinese-Alpaca-33B &\cellcolor[HTML]{dddbd4}0.326 &\cellcolor[HTML]{dcdad4}0.368 &\cellcolor[HTML]{d3d1cd}0.300 &\cellcolor[HTML]{e3e1d9}0.314 &\cellcolor[HTML]{e4e2d9}0.713 &\cellcolor[HTML]{e1dfd8}0.428 &\cellcolor[HTML]{e0ded7}0.288 &\cellcolor[HTML]{dcdbd4}0.303 &\cellcolor[HTML]{e2e0d8}0.295 &\cellcolor[HTML]{d7d5d0}0.401 &\cellcolor[HTML]{e4e2d9}0.199 \\
Tigerbot-7B-Chat &\cellcolor[HTML]{dcdad4}0.325 &\cellcolor[HTML]{d6d4d0}0.355 &\cellcolor[HTML]{d8d7d1}0.313 &\cellcolor[HTML]{dcdbd4}0.292 &\cellcolor[HTML]{e4e2d9}0.713 &\cellcolor[HTML]{dedcd5}0.415 &\cellcolor[HTML]{dfddd6}0.283 &\cellcolor[HTML]{e2e0d8}0.315 &\cellcolor[HTML]{e1dfd7}0.290 &\cellcolor[HTML]{d3d2ce}0.389 &\cellcolor[HTML]{d5d3cf}0.171 \\
Chinese-Alpaca-2-7B &\cellcolor[HTML]{dbd9d3}0.323 &\cellcolor[HTML]{e0ded7}0.375 &\cellcolor[HTML]{dad9d3}0.318 &\cellcolor[HTML]{dcdad4}0.289 &\cellcolor[HTML]{dddbd4}0.698 &\cellcolor[HTML]{dedcd6}0.417 &\cellcolor[HTML]{dfddd6}0.285 &\cellcolor[HTML]{dcdad4}0.303 &\cellcolor[HTML]{e4e2d9}0.312 &\cellcolor[HTML]{e3e1d9}0.439 &\cellcolor[HTML]{e3e1d9}0.193 \\
Aquilachat-7B &\cellcolor[HTML]{d4d3ce}0.309 &\cellcolor[HTML]{cdccc9}0.337 &\cellcolor[HTML]{e3e1d9}0.342 &\cellcolor[HTML]{cccbc9}0.236 &\cellcolor[HTML]{b6b6b9}0.609 &\cellcolor[HTML]{b4b4b8}0.255 &\cellcolor[HTML]{d6d4cf}0.249 &\cellcolor[HTML]{f0efd9}0.400 &\cellcolor[HTML]{fbfada}0.527 &\cellcolor[HTML]{e1dfd7}0.430 &\cellcolor[HTML]{fbfada}0.306 \\
Moss-Moon-003-Sft &\cellcolor[HTML]{d1cfcc}0.302 &\cellcolor[HTML]{c3c2c2}0.317 &\cellcolor[HTML]{dcdad4}0.321 &\cellcolor[HTML]{d5d4cf}0.267 &\cellcolor[HTML]{dbdad3}0.694 &\cellcolor[HTML]{d3d2ce}0.375 &\cellcolor[HTML]{d6d5d0}0.251 &\cellcolor[HTML]{c6c5c5}0.259 &\cellcolor[HTML]{e1dfd7}0.288 &\cellcolor[HTML]{dfddd6}0.424 &\cellcolor[HTML]{c7c6c5}0.152 \\
Qwen-7B-Chat &\cellcolor[HTML]{d0cfcc}0.301 &\cellcolor[HTML]{c7c6c5}0.325 &\cellcolor[HTML]{d2d0cd}0.297 &\cellcolor[HTML]{d9d7d1}0.278 &\cellcolor[HTML]{d6d4d0}0.681 &\cellcolor[HTML]{dfddd6}0.419 &\cellcolor[HTML]{dad9d3}0.266 &\cellcolor[HTML]{c2c1c2}0.251 &\cellcolor[HTML]{d7d5d0}0.248 &\cellcolor[HTML]{cdccc9}0.371 &\cellcolor[HTML]{cac9c8}0.157 \\
Belle-13B-Sft &\cellcolor[HTML]{bebdbf}0.264 &\cellcolor[HTML]{c3c2c2}0.317 &\cellcolor[HTML]{cccbc9}0.284 &\cellcolor[HTML]{cecdca}0.242 &\cellcolor[HTML]{c0bfc0}0.631 &\cellcolor[HTML]{bfbec0}0.299 &\cellcolor[HTML]{d4d3cf}0.244 &\cellcolor[HTML]{b4b4b8}0.222 &\cellcolor[HTML]{d3d2ce}0.234 &\cellcolor[HTML]{b4b4b8}0.296 &\cellcolor[HTML]{bab9bc}0.133 \\
CPM-Bee-10B &\cellcolor[HTML]{b4b4b8}0.244 &\cellcolor[HTML]{b4b4b8}0.286 &\cellcolor[HTML]{b4b4b8}0.224 &\cellcolor[HTML]{b4b4b8}0.147 &\cellcolor[HTML]{b4b4b8}0.603 &\cellcolor[HTML]{b9b9bb}0.277 &\cellcolor[HTML]{b4b4b8}0.117 &\cellcolor[HTML]{c8c7c6}0.263 &\cellcolor[HTML]{d0cfcc}0.220 &\cellcolor[HTML]{c6c5c5}0.352 &\cellcolor[HTML]{b4b4b8}0.125 \\
\bottomrule
\end{tabular}
}
\end{table*}

\section{Experiments}

\paragraph{Baselines.} We compare the performance of existing LLMs that have been trained on Chinese corpora. We select ChatGPT, for which we use \texttt{gpt-3.5-turbo-instruct},\footnote{https://openai.com/} which we believe corresponds to instructGPT text-davinci-002. Then we select a series of open-source LLMs, including \texttt{ChatGLM} \citep{zeng2023glm-130b}, \texttt{AquilaChat-7B}.\footnote{https://github.com/FlagAI-Open/FlagAI/} \texttt{Baichuan} \citep{baichuan2023baichuan2}, \texttt{Deepseek-Llm-67B-Chat} \citep{deepseek-llm}, \texttt{Qwen} \citep{qwen}, \texttt{Yi},\footnote{https://github.com/OrionStarAI/OrionStar-Yi-34B-Chat/tree/main} \texttt{tigerbot-7b-chat} \citep{DBLP:journals/corr/abs-2312-08688}, \texttt{TeleChat} \citep{DBLP:journals/corr/abs-2401-03804}, \texttt{CPM-Bee-10B},\footnote{https://github.com/OpenBMB/CPM-Bee}, and \texttt{Moss-Moon} \citep{sun2023moss}, which have been trained from scratch on a large volume of data in both English and Chinese. We additionally select other instruction-following LLMs, such as \texttt{Ziya-LLaMA-13B} \citep{fengshenbang}, \texttt{Chinese-Alpaca} \citep{chinese-llama-alpaca}, \texttt{Linly-Chinese-LLaMA2} \citep{DBLP:conf/acl/ZhaoLHZTLCSLMGG23}, and \texttt{BELLE} \citep{BELLE}, which are trained with Supervised Fine-Tuning (SFT) on Chinese data, including web texts, books, and code, and then trained via alignment techniques.

\paragraph{Settings.} 
For inference, we use four Nvidia A100 GPUs with 80GB of VRAM.
To optimize GPU resource usage, we directly employed the vLLM framework~\citep{kwon2023efficient} for LLM inference on CIF-Bench where applicable. 
This setup enables each model to complete all tasks within approximately 6 to 12 hours. 
For models not supported by the vLLM, we adhere to the configurations specified in official repositories, resulting in an inference duration ranging from 12 to 48 hours.
During the evaluation, we use two Nvidia 2080-Ti 12GB GPUs to conduct the BLEURT semantic similarity calculations, and use the \texttt{gpt-4-turbo-preview} version of GPT-4 API as the open-ended evaluator for the rest of tasks.

\begin{table}[!htp]\centering
\caption{Comparison between English-translated and newly annotated Chinese tasks in the \textit{Public} split.}\label{tab:compare_public_sni_new_task}
\scriptsize
\begin{tabular}{lcc}\toprule
\textbf{Model} &\textbf{SNI Task} &\textbf{New Task} \\\midrule
Qwen-72B-Chat &0.588 &0.573 \\
Qwen-14B-Chat &0.573 &0.535 \\
Deepseek-LLM-67B-Chat &0.529 &0.504 \\
gpt-3.5-public-turbo &0.523 &0.500 \\
Yi-34B-Chat &0.509 &0.514 \\
\bottomrule
\end{tabular}
\end{table}

\begin{table}[!htbp]\centering
\caption{Comparison of the CIF-Bench overall scores in the \textit{Public} split and other leaderboards. The cells are highlighted with fading colors from \colorbox[HTML]{fbfada}{maximum} to \colorbox[HTML]{b4b4b8}{minimum} for the applicable numbers in a column. \textbf{*} indicates that the performance of pre-trained base LLMs is used to approximate the evaluation of the corresponding unavailable chat models.}\label{tab: compare CIF and others}
\scriptsize
\scalebox{0.85}{
\begin{tabular}{lrrrr}\toprule
\textbf{Model Name} &\textbf{CIF \textit{Public}} &\textbf{Open LLM} &\textbf{OpenCompass} \\\midrule
Qwen-72B-Chat &\cellcolor[HTML]{fbfada}0.589 &\cellcolor[HTML]{fbfada}*73.60 &\cellcolor[HTML]{fbfada}51.90 \\
Qwen-14B-Chat &\cellcolor[HTML]{f6f4d9}0.564 &\cellcolor[HTML]{f1efd9}*65.86 &\cellcolor[HTML]{edebd9}45.00 \\
Deepseek-LLM-67B-Chat &\cellcolor[HTML]{eeedd9}0.526 &\cellcolor[HTML]{f8f7d9}71.79 &\cellcolor[HTML]{e8e6d9}42.70 \\
gpt-3.5-Public-SFT &\cellcolor[HTML]{edecd9}0.522 &- &\cellcolor[HTML]{f0efd9}46.80 \\
Yi-34B-Chat &\cellcolor[HTML]{ecebd9}0.516 &\cellcolor[HTML]{f0efd9}65.32 &\cellcolor[HTML]{f1f0d9}47.10 \\
Baichuan2-13B-Chat &\cellcolor[HTML]{ebead9}0.512 &- &\cellcolor[HTML]{c0bfc0}32.10 \\
Tigerbot-13B-Chat &\cellcolor[HTML]{e8e6d9}0.494 &\cellcolor[HTML]{e0ded7}*53.42 &- \\
Chinese-Alpaca-2-13B &\cellcolor[HTML]{e7e6d9}0.492 &\cellcolor[HTML]{e6e4d9}57.41 &- \\
Chinese-Alpaca-33B &\cellcolor[HTML]{e6e4d9}0.484 &\cellcolor[HTML]{e3e1d9}55.33 &- \\
Ziya-Llama-13B &\cellcolor[HTML]{e5e3d9}0.479 &\cellcolor[HTML]{b4b4b8}29.96 &- \\
Chinese-Llama2-Linly-13B &\cellcolor[HTML]{e5e3d9}0.479 &- &- \\
Tigerbot-7B-Chat &\cellcolor[HTML]{e5e3d9}0.478 &\cellcolor[HTML]{d6d4cf}*47.93 &- \\
ChatGLM3-6B &\cellcolor[HTML]{e3e1d9}0.472 &- &\cellcolor[HTML]{cdccca}35.20 \\
Chinese-Alpaca-13B &\cellcolor[HTML]{e3e1d9}0.471 &- &- \\
ChatGLM2-6B &\cellcolor[HTML]{e1dfd8}0.464 &- &- \\
Chinese-Alpaca-7B &\cellcolor[HTML]{dddbd4}0.452 &\cellcolor[HTML]{d7d6d1}48.85 &- \\
Chinese-Alpaca-2-7B &\cellcolor[HTML]{dbdad4}0.448 &- &- \\
Chinese-Llama2-Linly-7B &\cellcolor[HTML]{d9d8d2}0.443 &\cellcolor[HTML]{d1d0cc}45.44 &- \\
Qwen-7B-Chat &\cellcolor[HTML]{d9d7d2}0.442 &\cellcolor[HTML]{e8e6d9}*59.19 &\cellcolor[HTML]{d6d4d0}37.10 \\
ChatGLM-6B &\cellcolor[HTML]{d8d7d1}0.440 &- &- \\
Baichuan-13B-Chat &\cellcolor[HTML]{d3d2ce}0.426 &- &- \\
Yi-6B-Chat &\cellcolor[HTML]{d1d0cc}0.420 &\cellcolor[HTML]{e1dfd8}*54.08 &\cellcolor[HTML]{bfbebf}31.90 \\
CPM-Bee-10B &\cellcolor[HTML]{cfcecb}0.415 &- &- \\
Moss-Moon-003-SFT &\cellcolor[HTML]{c9c8c7}0.399 &- &- \\
Belle-SFT-Public &\cellcolor[HTML]{c9c8c7}0.397 &- &- \\
Telechat-7B &\cellcolor[HTML]{b8b7ba}0.350 &- &- \\
Aquilachat-7B &\cellcolor[HTML]{b7b7ba}0.350 &- &- \\
Baichuan2-7B-Chat &\cellcolor[HTML]{b4b4b8}0.339 &\cellcolor[HTML]{dcdbd4}51.42 &\cellcolor[HTML]{b4b4b8}29.40 \\
\bottomrule
\end{tabular}
}
\end{table}

\begin{table*}[!htp]\centering
\caption{Overall performance differences in CIF-Bench from \textit{Public} to \textit{Private} splits with single instructions.}\label{tab: compare private-public}
\scriptsize
\begin{tabular}{lr|lr}\toprule
\textbf{Model Name} &\textbf{Score Difference$\uparrow$} &\textbf{Model Name} &\textbf{Score Difference$\uparrow$} \\\midrule
Aquilachat-7B &\textcolor[HTML]{990000}{-0.050$\downarrow$} &Chinese-Llama2-Linly-7B &\textcolor[HTML]{990000}{-0.122$\downarrow$} \\
Baichuan-13B-Chat &\textcolor[HTML]{38761d}{0.020$\uparrow$} &CPM-Bee-10B &\textcolor[HTML]{990000}{-0.178$\downarrow$} \\
Baichuan2-13B-Chat &\textcolor[HTML]{38761d}{0.006$\uparrow$} &Deepseek-LLM-67B-Chat &\textcolor[HTML]{990000}{-0.060$\downarrow$} \\
Baichuan2-7B-Chat &\textcolor[HTML]{38761d}{0.071$\uparrow$} &gpt-3.5-Public-SFT &\textcolor[HTML]{990000}{-0.187$\downarrow$} \\
Belle-SFT-Public &\textcolor[HTML]{990000}{-0.145$\downarrow$} &Moss-Moon-003-SFT &\textcolor[HTML]{990000}{-0.110$\downarrow$} \\
ChatGLM-6B &\textcolor[HTML]{990000}{-0.112$\downarrow$} &Qwen-14B-Chat &\textcolor[HTML]{990000}{-0.068$\downarrow$} \\
ChatGLM2-6B &\textcolor[HTML]{990000}{-0.124$\downarrow$} &Qwen-72B-Chat &\textcolor[HTML]{990000}{-0.068$\downarrow$} \\
ChatGLM3-6B &\textcolor[HTML]{990000}{-0.038$\downarrow$} &Qwen-7B-Chat &\textcolor[HTML]{990000}{-0.145$\downarrow$} \\
Chinese-Alpaca-13B &\textcolor[HTML]{990000}{-0.148$\downarrow$} &Telechat-7B &\textcolor[HTML]{990000}{-0.029$\downarrow$} \\
Chinese-Alpaca-2-13B &\textcolor[HTML]{990000}{-0.171$\downarrow$} &Tigerbot-13B-Chat &\textcolor[HTML]{990000}{-0.180$\downarrow$} \\
Chinese-Alpaca-2-7B &\textcolor[HTML]{990000}{-0.138$\downarrow$} &Tigerbot-7B-Chat &\textcolor[HTML]{990000}{-0.163$\downarrow$} \\
Chinese-Alpaca-33B &\textcolor[HTML]{990000}{-0.170$\downarrow$} &Yi-34B-Chat &\textcolor[HTML]{990000}{-0.014$\downarrow$} \\
Chinese-Alpaca-7B &\textcolor[HTML]{990000}{-0.125$\downarrow$} &Yi-6B-Chat &\textcolor[HTML]{990000}{-0.008$\downarrow$} \\
Chinese-Llama2-Linly-13B &\textcolor[HTML]{990000}{-0.147$\downarrow$} &Ziya-Llama-13B &\textcolor[HTML]{990000}{-0.167$\downarrow$} \\
\bottomrule
\end{tabular}
\end{table*}

\begin{table*}[!htp]\centering
\caption{The performance shift caused by unseen data instances and unseen tasks. 
Note that in the column ``Existing'' task, only the newly annotated and existing input-output data instances are compared while the task instruction remains the same.
In the ``Existing$\rightarrow$New'' setting, both data instances and tasks are changed.
}
\label{tab:new_task_instance_degrade}
\scriptsize
\begin{tabular}{lr|rr|l|rr}\toprule
\textbf{Model $\downarrow$} &\textbf{Task$\rightarrow$}&\textbf{Existing} &\textbf{Existing$\rightarrow$New} &\textbf{Model$\downarrow$} &\textbf{Existing} &\textbf{Existing$\rightarrow$New} \\\midrule
\multicolumn{2}{l|}{Aquilachat-7B} &\textcolor[HTML]{990000}{-0.047$\downarrow$} &\textcolor[HTML]{990000}{-0.034$\downarrow$} &Chinese-Llama2-Linly-7B&\textcolor[HTML]{990000}{-0.134$\downarrow$} &\textcolor[HTML]{990000}{-0.047$\downarrow$} \\
\multicolumn{2}{l|}{Baichuan-13B-Chat} & \textcolor[HTML]{38761d}{0.023$\uparrow$} & \textcolor[HTML]{38761d}{0.027$\uparrow$} & CPM-Bee-10B &\textcolor[HTML]{990000}{-0.176$\downarrow$} & \textcolor[HTML]{38761d}{0.046$\uparrow$} \\
\multicolumn{2}{l|}{Baichuan2-13B-Chat} &\textcolor[HTML]{990000}{-0.003$\downarrow$} &\textcolor[HTML]{38761d}{0.008$\uparrow$} &Deepseek-LLM-67B-Chat &\textcolor[HTML]{990000}{-0.076$\downarrow$} &\textcolor[HTML]{990000}{-0.029$\downarrow$} \\
\multicolumn{2}{l|}{Baichuan2-7B-Chat} &\textcolor[HTML]{38761d}{0.072$\uparrow$} & \textcolor[HTML]{38761d}{0.077$\uparrow$} &gpt-3.5-Public-SFT &\textcolor[HTML]{990000}{-0.202$\downarrow$} &\textcolor[HTML]{990000}{-0.029$\downarrow$} \\
\multicolumn{2}{l|}{Belle-SFT-Public} &\textcolor[HTML]{990000}{-0.167$\downarrow$} &\textcolor[HTML]{990000}{-0.054$\downarrow$} &Moss-Moon-003-SFT &\textcolor[HTML]{990000}{-0.124$\downarrow$} &\textcolor[HTML]{990000}{-0.028$\downarrow$} \\
\multicolumn{2}{l|}{ChatGLM-6B} &\textcolor[HTML]{990000}{-0.120$\downarrow$} &\textcolor[HTML]{990000}{-0.033$\downarrow$} &Qwen-14B-Chat &\textcolor[HTML]{990000}{-0.088$\downarrow$} &\textcolor[HTML]{990000}{-0.038$\downarrow$} \\
\multicolumn{2}{l|}{ChatGLM2-6B} &\textcolor[HTML]{990000}{-0.131$\downarrow$} &\textcolor[HTML]{990000}{-0.005$\downarrow$} &Qwen-72B-Chat &\textcolor[HTML]{990000}{-0.082$\downarrow$} &\textcolor[HTML]{990000}{-0.021$\downarrow$} \\
\multicolumn{2}{l|}{ChatGLM3-6B} &\textcolor[HTML]{990000}{-0.060$\downarrow$} &\textcolor[HTML]{990000}{-0.052$\downarrow$} &Qwen-7B-Chat &\textcolor[HTML]{990000}{-0.157$\downarrow$} &\textcolor[HTML]{38761d}{0.005$\uparrow$} \\
\multicolumn{2}{l|}{Chinese-Alpaca-13B} &\textcolor[HTML]{990000}{-0.164$\downarrow$} &\textcolor[HTML]{990000}{-0.081$\downarrow$} &Telechat-7B &\textcolor[HTML]{990000}{-0.050$\downarrow$} &\textcolor[HTML]{990000}{-0.045$\downarrow$} \\
\multicolumn{2}{l|}{Chinese-Alpaca-2-13B} &\textcolor[HTML]{990000}{-0.179$\downarrow$} &\textcolor[HTML]{990000}{-0.067$\downarrow$} &Tigerbot-13B-Chat &\textcolor[HTML]{990000}{-0.187$\downarrow$} &\textcolor[HTML]{990000}{-0.017$\downarrow$} \\
\multicolumn{2}{l|}{Chinese-Alpaca-2-7B} &\textcolor[HTML]{990000}{-0.152$\downarrow$} &\textcolor[HTML]{990000}{-0.051$\downarrow$} &Tigerbot-7B-Chat &\textcolor[HTML]{990000}{-0.162$\downarrow$} &\textcolor[HTML]{990000}{-0.004$\downarrow$} \\
\multicolumn{2}{l|}{Chinese-Alpaca-33B} &\textcolor[HTML]{990000}{-0.187$\downarrow$} &\textcolor[HTML]{990000}{-0.072$\downarrow$} &Yi-34B-Chat &\textcolor[HTML]{990000}{-0.025$\downarrow$} &\textcolor[HTML]{990000}{-0.002$\downarrow$} \\
\multicolumn{2}{l|}{Chinese-Alpaca-7B} &\textcolor[HTML]{990000}{-0.147$\downarrow$} &\textcolor[HTML]{990000}{-0.072$\downarrow$} &Yi-6B-Chat &\textcolor[HTML]{990000}{-0.022$\downarrow$} &\textcolor[HTML]{990000}{-0.012$\downarrow$} \\
\multicolumn{2}{l|}{Chinese-Llama2-Linly-13B} &\textcolor[HTML]{990000}{-0.153$\downarrow$} &\textcolor[HTML]{990000}{-0.031$\downarrow$} &Ziya-Llama-13B &\textcolor[HTML]{990000}{-0.181$\downarrow$} &\textcolor[HTML]{990000}{-0.045$\downarrow$} \\
\bottomrule
\end{tabular}
\end{table*}

\section{Results Analysis}

Broadly speaking, we aim to investigate the performance capabilities of current representative Chinese LLMs in a diverse set of NLP tasks to ascertain how well the annotated data with human standards with the provided instruction-following benchmark. 
Specifically, we ask: 
\emph{(i)} Is our benchmark challenging enough? What kind of tasks are difficult?
\emph{(ii)} Is it true that LLMs perform worse when language is transferred?
\emph{(iii)} Do we measure the instruction-following capability well, by avoiding data contamination? 
\emph{(iv)} Do the diverse instructions help?

\paragraph{Is CIF-Bench Challenging?}

To ensure the reliability of our benchmark, the scores in the \textit{private} split with the diversified instructions are referred to as the main results for discussion, as shown in \autoref{tab: overall result 1 private} and \autoref{tab: overall result 2 private}.
Our findings reveal that although large parameter size contributes to performance (\texttt{Qwen-72B-Chat}, \texttt{Yi-34B-Chat}, and \texttt{Deepseek-LLM-67B-Chat}), the effective training methods are still a boost for relatively small models such as \texttt{Baichuan2-13B-Chat} and \texttt{Qwen-14B-Chat}.
Given that the highest score barely reaches $52.9$ overall out of $100$ and only 4 models exceed $50.0$, we conclude that our proposed CIF is a tough benchmark for existing LLMs for question \emph{(i)}.

In addition, we provide finer-grained score aggregation to further analyze the challenging task categories (n.b., most bilingual LLMs perform poorly on tasks in code, summarization, and translation categories).
In the code category, the models might misunderstand the semantics expressed in Chinese for the newly defined variable or function. Specifically, models usually perform poorly in a new ``programming language'' environment that requires the model to understand restricted actions.
As for summarization tasks, models could misinterpret the instruction, eg. models sometimes consider the instruction ``modify the input into a more friendly expression to non-native speakers'' as a Chinese-English translation task and might provide redundant explanations even if not required by the instructions and hence will cause large semantic distances to the golden reference.
We point out that Chinese-commented code corpora and parallel translation data of Chinese and other languages are still scarce resources, which might lead to their poor performance on CIF-Bench's code and translation categories.
Additionally, we assume that Chinese and English bilingual LLMs, although a major branch of multilingual LLM, do not significantly benefit LLMs' capacity to deal with minor-language-related tasks.
Part of the tasks in CIF-Bench's summarization category are very challenging, combining counterfactual reasoning and empathy estimation (i.e., task 125 and task 131 referring to Appendix~\ref{apdx:tasks}).
Thereby, the bilingual LLMs' poor performance on CIF-Bench's summarization category is understandable.
Detailed category-based scores on the \textit{public} split are available in \autoref{tab: overall result 1 public} in Appendix~\ref{apdx:cif_public_scores} for further analysis.

\paragraph{Language Transferability.}

We select the \textit{public} split to investigate LLM language transferability in instruction-following.
In the CIF-Bench \textit{public} split, a set of 70 tasks from SNI~\cite{wang2022super} are used as representative samples of English NLP tasks equipped with directly translated input-output pairs in Chinese.
We select the top-5 performing models on the \textit{public} split to show the performance comparison between SNI and our $37$ original curated Chinese tasks in \autoref{tab:compare_public_sni_new_task}.
Although these models maintain instruction-following capability when encountering the translated SNI data, they generally perform worse on tasks newly created in Chinese without a corresponding ``copy'' in English, which yields an average score decrement of $2.2\%$.

\begin{table*}[!htp]\centering
\caption{The difference of variance of task-level scores from single to diverse instruction sets. The variance values are scaled by a factor of $1 \times 10^{-3}$.}\label{tab: diverse_instruction_var_change}
\scriptsize
\begin{tabular}{lc|lc}\toprule
\textbf{Model Name} &\textbf{Var. Difference$\downarrow$} &\textbf{Model Name} &\textbf{Var. Difference$\downarrow$} \\\midrule
Aquilachat-7B &\textcolor[HTML]{38761d}{-3.961$\downarrow$} &Chinese-Llama2-Linly-7B &\textcolor[HTML]{38761d}{-4.539$\downarrow$} \\
Baichuan-13B-Chat &\textcolor[HTML]{38761d}{-7.049$\downarrow$} &Cpm-Bee-10B &\textcolor[HTML]{38761d}{-0.661$\downarrow$} \\
Baichuan2-13B-Chat &\textcolor[HTML]{38761d}{-3.633$\downarrow$} &Deepseek-Llm-67B-Chat &\textcolor[HTML]{38761d}{-2.889$\downarrow$} \\
Baichuan2-7B-Chat &\textcolor[HTML]{38761d}{-1.402$\downarrow$} &Gpt-3.5-Turbo-Sft &\textcolor[HTML]{38761d}{-6.369$\downarrow$} \\
Belle-13B-Sft &\textcolor[HTML]{38761d}{-0.316$\downarrow$} &Moss-Moon-003-Sft &\textcolor[HTML]{38761d}{-6.827$\downarrow$} \\
Chatglm-6B-Sft &\textcolor[HTML]{38761d}{-5.051$\downarrow$} &Qwen-14B-Chat &\textcolor[HTML]{38761d}{-0.978$\downarrow$} \\
Chatglm2-6B &\textcolor[HTML]{38761d}{-3.980$\downarrow$} &Qwen-72B-Chat &\textcolor[HTML]{38761d}{-1.817$\downarrow$} \\
Chatglm3-6B &\textcolor[HTML]{38761d}{-0.413$\downarrow$} &Qwen-7B-Chat &\textcolor[HTML]{38761d}{-3.185$\downarrow$} \\
Chinese-Alpaca-13B &\textcolor[HTML]{38761d}{-8.303$\downarrow$} &Telechat-7B &\textcolor[HTML]{38761d}{-6.090$\downarrow$} \\
Chinese-Alpaca-2-13B &\textcolor[HTML]{38761d}{-4.814$\downarrow$} &Tigerbot-13B-Chat &\textcolor[HTML]{38761d}{-3.816$\downarrow$} \\
Chinese-Alpaca-2-7B &\textcolor[HTML]{38761d}{-4.494$\downarrow$} &Tigerbot-7B-Chat &\textcolor[HTML]{38761d}{-6.004$\downarrow$} \\
Chinese-Alpaca-33B &\textcolor[HTML]{38761d}{-5.000$\downarrow$} &Yi-34B-Chat &\textcolor[HTML]{38761d}{-1.942$\downarrow$} \\
Chinese-Alpaca-7B &\textcolor[HTML]{38761d}{-2.961$\downarrow$} &Yi-6B-Chat &\textcolor[HTML]{38761d}{-6.397$\downarrow$} \\
Chinese-Llama2-Linly-13B &\textcolor[HTML]{38761d}{-2.961$\downarrow$} &Ziya-Llama-13B &\textcolor[HTML]{38761d}{-3.001$\downarrow$} \\
\bottomrule
\end{tabular}
\end{table*}

\paragraph{Data Contamination Does Exist.}\label{sec:private_gap}

As mentioned in \S\ref{sec:benchmark}, we evaluate the model performances on the \textit{public} split with half of the input-output pairs in the single instruction setting, with which we can conveniently probe the benchmark data contamination issue of the LLMs.

We first compare the CIF-Bench \textit{public} results with two comprehensive LLM benchmarks, including the Open LLM Leaderboard~\cite{open-llm-leaderboard}, as well as an English-Chinese leaderboard, OpenCompass~\cite{2023opencompass}. 
As suggested in \autoref{tab: compare CIF and others} with rows ranked in the descending order of the overall \textit{public} scores, the results on CIF-Bench are aligned with the other two popular benchmarks, which therefore verifies the reliability of our evaluation pipeline.
However, we suspect the highly correlative rankings could be a result of the benchmark data leakage in those ``web-scale'' pre-training data, since $117$ of the constructed tasks and instances in the \textit{public} split are sourced from the internet.

To further confirm such suspicions, we calculate the performance changes of overall scores in the same single instruction setting, but with different input-output pairs from the \textit{public} and \textit{private} splits. Revealed by the differences in \autoref{tab: compare private-public}, there is a noticeable performance drop for most (25/28) of the models when a large part of the data translated from public sources is replaced by our original annotations.
Consequently, incoming models submitted to the proposed CIF-Bench will restricted to the \textit{private} split for the sake of evaluation reliability.

It is likely that both the leakage of the input-output instances and the tasks themselves contribute to the mentioned evaluation bias.
To compare the two factors for the downgraded performances, 
we analyze the performance shift with the $113$ ``Existing'' tasks translated from English or originally in Chinese and the $37$ ``New'' tasks we crafted from scratch. 
As revealed in \autoref{tab:new_task_instance_degrade}, the LLMs have impaired performance when given newly curated data instances for a set of seen ``Existing'' tasks, yielding an average $11.0\%$ score decrease. In contrast, these models on average perform $2.5\%$ worse, with both definitely-unseen tasks and corresponding input-output pairs.
We hence conclude that the leakage of the data instances plays a more significant role than the tasks themselves in evaluation biases.

\paragraph{Instruction Diversity for Evaluation Robustness.}\label{sec:instruction_diversity}

With the motivation that a model might produce inconsistent \ul{output} given various \ul{instruction},\ul{input} holding the same semantics, we argue that a diversified instruction set can increase the evaluation robustness by incorporating more corner cases.
We separately calculate the task-level score variance in the \textit{private} split for the conditions of using \textit{one} and \textit{five} instructions to verify the improvement.
We find that increasing the diversity of the task instructions can bring extra robustness to the evaluation, as the evaluation scores are stabilized to lower variance for all the tested LLMs (see in Table~\ref{tab: diverse_instruction_var_change}).

\paragraph{Human Annotation for Verification.}

To verify the annotation quality and reliability , we invite 3 annotators with expert-level NLP research backgrounds to assess the model outputs in \textit{public} split with the same task-level instruction. 
The evaluation dimensions include: ``Faithfulness'': human experts reflect on the absolute quality of a model's output in a binary (yes/no) form. ``Level of preference'': a 5-point Likert scale was provided to the experts to assess the relative quality of the model outputs. 
We randomly sample tasks according to the task category distribution,
and pick three models performing differently in general, specifically \texttt{Moss-Moon-003-sft} ($0.399$), \texttt{Baichuan-13B-Chat} ($0.426$), and \texttt{Qwen-72B-Chat} ($0.589$).
Considering the diverse and open-ended task, we first measure quality by comparing the pairwise agreement between two annotators, reporting an average agreement of 0.49. Furthermore, we employ Cohen's kappa \citep{DBLP:journals/eswa/Ben-David08} to measure inter-rater reliability, reporting an average of 0.3729 across the $153$ questions, implying that the results are substantially reliable. 
Specifically, the experts scored $0.4966$ on the dichotomous form yet $0.2492$ on the more varied options, suggesting that completing $153$ questions is challenging even for human experts. We further explore the correlation between the model prediction with human evaluation(Spearman's $r=0.4043$), suggesting that most annotated were indeed truthful and the models can be relied upon to generate output for this task.

\section{Conclusion}
In summary, CIF-Bench not only exposes the limitations of current LLMs in navigating the complexities of Chinese language instruction-following tasks but also provides a foundational platform for future advancements in LLM generalizability research. 
Through this work, we aim to facilitate the development of more adaptable, culturally aware, and linguistically diverse language models, capable of truly understanding and interacting with the global tapestry of human language.

\section*{Limitations}
Recruiting human subjects for annotation limits the reproducibility of human evaluation. In addition, we recognize that there might be more suitable baseline models, whilst in this study only a few of the most advanced models were used. Finally, despite annotation and discrimination by human experts, there may still be offensive content in the data due to both human education and environmental factors. It is worth noting, however, that identifying offensive language is not the purpose of this work.

\section*{Ethics Statement}
The dataset presented was annotated by a third-party professional annotation company. During the annotation process, we considered the following aspects to ensure the protection of the annotators. (1) Consent: To ensure that our participants agreed to the annotation task, we asked them to read the task guidelines and instructions before starting the work. If they felt uncomfortable, they could withdraw from the task at any time. (2) Confidentiality: The entire annotation process was anonymous and we did not know any information about the participants in the task. (3) Assurance: all data were obtained from open-source datasets or resources.

\section*{Acknowledgements}
Yizhi Li and Xingwei Qu are Ph.D. students funded by the Department of Computer Science,
University of Manchester, UK.

\bibliography{anthology}

\begin{thebibliography}{66}
\expandafter\ifx\csname natexlab\endcsname\relax\def\natexlab#1{#1}\fi

\bibitem[{Altammami et~al.(2020)Altammami, Atwell, and Alsalka}]{altammami-etal-2020-constructing}
Shatha Altammami, Eric Atwell, and Ammar Alsalka. 2020.
\newblock \href {https://aclanthology.org/2020.lrec-1.415} {Constructing a bilingual {H}adith corpus using a segmentation tool}.
\newblock In \emph{Proceedings of the Twelfth Language Resources and Evaluation Conference}, pages 3390--3398, Marseille, France. European Language Resources Association.

\bibitem[{Bach et~al.(2022)Bach, Sanh, Yong, Webson, Raffel, Nayak, Sharma, Kim, Bari, Fevry et~al.}]{bach2022promptsource}
Stephen~H Bach, Victor Sanh, Zheng-Xin Yong, Albert Webson, Colin Raffel, Nihal~V Nayak, Abheesht Sharma, Taewoon Kim, M~Saiful Bari, Thibault Fevry, et~al. 2022.
\newblock Promptsource: An integrated development environment and repository for natural language prompts.
\newblock \emph{arXiv preprint arXiv:2202.01279}.

\bibitem[{Bai et~al.(2024)Bai, Liu, Bu, He, Liu, Zhou, Lin, Su, Ge, Zheng, and Ouyang}]{bai2024mtbench101}
Ge~Bai, Jie Liu, Xingyuan Bu, Yancheng He, Jiaheng Liu, Zhanhui Zhou, Zhuoran Lin, Wenbo Su, Tiezheng Ge, Bo~Zheng, and Wanli Ouyang. 2024.
\newblock \href {http://arxiv.org/abs/2402.14762} {Mt-bench-101: A fine-grained benchmark for evaluating large language models in multi-turn dialogues}.

\bibitem[{Bai et~al.(2023)Bai, Bai, Chu, Cui, Dang, Deng, Fan, Ge, Han, Huang, Hui, Ji, Li, Lin, Lin, Liu, Liu, Lu, Lu, Ma, Men, Ren, Ren, Tan, Tan, Tu, Wang, Wang, Wang, Wu, Xu, Xu, Yang, Yang, Yang, Yang, Yao, Yu, Yuan, Yuan, Zhang, Zhang, Zhang, Zhang, Zhou, Zhou, Zhou, and Zhu}]{qwen}
Jinze Bai, Shuai Bai, Yunfei Chu, Zeyu Cui, Kai Dang, Xiaodong Deng, Yang Fan, Wenbin Ge, Yu~Han, Fei Huang, Binyuan Hui, Luo Ji, Mei Li, Junyang Lin, Runji Lin, Dayiheng Liu, Gao Liu, Chengqiang Lu, Keming Lu, Jianxin Ma, Rui Men, Xingzhang Ren, Xuancheng Ren, Chuanqi Tan, Sinan Tan, Jianhong Tu, Peng Wang, Shijie Wang, Wei Wang, Shengguang Wu, Benfeng Xu, Jin Xu, An~Yang, Hao Yang, Jian Yang, Shusheng Yang, Yang Yao, Bowen Yu, Hongyi Yuan, Zheng Yuan, Jianwei Zhang, Xingxuan Zhang, Yichang Zhang, Zhenru Zhang, Chang Zhou, Jingren Zhou, Xiaohuan Zhou, and Tianhang Zhu. 2023.
\newblock Qwen technical report.
\newblock \emph{arXiv preprint arXiv:2309.16609}.

\bibitem[{Baichuan(2023)}]{baichuan2023baichuan2}
Baichuan. 2023.
\newblock \href {https://arxiv.org/abs/2309.10305} {Baichuan 2: Open large-scale language models}.
\newblock \emph{arXiv preprint arXiv:2309.10305}.

\bibitem[{Bara et~al.(2021)Bara, CH-Wang, and Chai}]{nt_bara-etal-2021-mindcraft}
Cristian-Paul Bara, Sky CH-Wang, and Joyce Chai. 2021.
\newblock \href {https://doi.org/10.18653/v1/2021.emnlp-main.85} {{M}ind{C}raft: Theory of mind modeling for situated dialogue in collaborative tasks}.
\newblock In \emph{Proceedings of the 2021 Conference on Empirical Methods in Natural Language Processing}, pages 1112--1125, Online and Punta Cana, Dominican Republic. Association for Computational Linguistics.

\bibitem[{Bawden et~al.(2021)Bawden, Bilinski, Lavergne, and Rosset}]{bawden_DiaBLa-A-Corpus-of_2021}
Rachel Bawden, Eric Bilinski, Thomas Lavergne, and Sophie Rosset. 2021.
\newblock \href {https://doi.org/10.1007/s10579-020-09514-4} {Diabla: A corpus of bilingual spontaneous written dialogues for machine translation}.
\newblock \emph{Language Resources and Evaluation}, 55:635--660.

\bibitem[{Beeching et~al.(2023)Beeching, Fourrier, Habib, Han, Lambert, Rajani, Sanseviero, Tunstall, and Wolf}]{open-llm-leaderboard}
Edward Beeching, Clémentine Fourrier, Nathan Habib, Sheon Han, Nathan Lambert, Nazneen Rajani, Omar Sanseviero, Lewis Tunstall, and Thomas Wolf. 2023.
\newblock Open llm leaderboard.
\newblock \url{https://huggingface.co/spaces/HuggingFaceH4/open_llm_leaderboard}.

\bibitem[{BELLEGroup(2023)}]{BELLE}
BELLEGroup. 2023.
\newblock Belle: Be everyone's large language model engine.
\newblock \url{https://github.com/LianjiaTech/BELLE}.

\bibitem[{Ben{-}David(2008)}]{DBLP:journals/eswa/Ben-David08}
Arie Ben{-}David. 2008.
\newblock Comparison of classification accuracy using cohen's weighted kappa.
\newblock \emph{Expert Syst. Appl.}, 34(2):825--832.

\bibitem[{bench authors(2023)}]{nt_srivastava2023beyond}
BIG bench authors. 2023.
\newblock \href {https://openreview.net/forum?id=uyTL5Bvosj} {Beyond the imitation game: Quantifying and extrapolating the capabilities of language models}.
\newblock \emph{Transactions on Machine Learning Research}.

\bibitem[{Chen et~al.(2023)Chen, Cai, Wu, Li, Xin, and Fu}]{DBLP:journals/corr/abs-2312-08688}
Ye~Chen, Wei Cai, Liangmin Wu, Xiaowei Li, Zhanxuan Xin, and Cong Fu. 2023.
\newblock Tigerbot: An open multilingual multitask {LLM}.
\newblock \emph{CoRR}, abs/2312.08688.

\bibitem[{Cobbe et~al.(2021)Cobbe, Kosaraju, Bavarian, Chen, Jun, Kaiser, Plappert, Tworek, Hilton, Nakano, Hesse, and Schulman}]{cobbe2021gsm8k}
Karl Cobbe, Vineet Kosaraju, Mohammad Bavarian, Mark Chen, Heewoo Jun, Lukasz Kaiser, Matthias Plappert, Jerry Tworek, Jacob Hilton, Reiichiro Nakano, Christopher Hesse, and John Schulman. 2021.
\newblock Training verifiers to solve math word problems.
\newblock \emph{arXiv preprint arXiv:2110.14168}.

\bibitem[{Contributors(2023)}]{2023opencompass}
OpenCompass Contributors. 2023.
\newblock Opencompass: A universal evaluation platform for foundation models.
\newblock \url{https://github.com/open-compass/opencompass}.

\bibitem[{C\^ot\'e et~al.(2018)C\^ot\'e, K\'ad\'ar, Yuan, Kybartas, Barnes, Fine, Moore, Tao, Hausknecht, Asri, Adada, Tay, and Trischler}]{nt_cote18textworld}
Marc-Alexandre C\^ot\'e, \'Akos K\'ad\'ar, Xingdi Yuan, Ben Kybartas, Tavian Barnes, Emery Fine, James Moore, Ruo~Yu Tao, Matthew Hausknecht, Layla~El Asri, Mahmoud Adada, Wendy Tay, and Adam Trischler. 2018.
\newblock Textworld: A learning environment for text-based games.
\newblock \emph{CoRR}, abs/1806.11532.

\bibitem[{Cui et~al.(2023)Cui, Yang, and Yao}]{chinese-llama-alpaca}
Yiming Cui, Ziqing Yang, and Xin Yao. 2023.
\newblock \href {https://arxiv.org/abs/2304.08177} {Efficient and effective text encoding for chinese llama and alpaca}.
\newblock \emph{arXiv preprint arXiv:2304.08177}.

\bibitem[{DeepSeek-AI(2024)}]{deepseek-llm}
DeepSeek-AI. 2024.
\newblock \href {https://github.com/deepseek-ai/DeepSeek-LLM} {Deepseek llm: Scaling open-source language models with longtermism}.
\newblock \emph{arXiv preprint arXiv:2401.02954}.

\bibitem[{Dua et~al.(2019)Dua, Wang, Dasigi, Stanovsky, Singh, and Gardner}]{nt_dua2019drop}
Dheeru Dua, Yizhong Wang, Pradeep Dasigi, Gabriel Stanovsky, Sameer Singh, and Matt Gardner. 2019.
\newblock Drop: A reading comprehension benchmark requiring discrete reasoning over paragraphs.
\newblock \emph{arXiv preprint arXiv:1903.00161}.

\bibitem[{Emelin et~al.(2021)Emelin, Le~Bras, Hwang, Forbes, and Choi}]{nt_emelin-etal-2021-moral}
Denis Emelin, Ronan Le~Bras, Jena~D. Hwang, Maxwell Forbes, and Yejin Choi. 2021.
\newblock \href {https://doi.org/10.18653/v1/2021.emnlp-main.54} {Moral stories: Situated reasoning about norms, intents, actions, and their consequences}.
\newblock In \emph{Proceedings of the 2021 Conference on Empirical Methods in Natural Language Processing}, pages 698--718, Online and Punta Cana, Dominican Republic. Association for Computational Linguistics.

\bibitem[{"European~Commission and Technology."(2017)}]{spanishen2017}
Content "European~Commission, Directorate-General for Communications~Networks and Technology.". 2017.
\newblock \href {http://data.europa.eu/88u/dataset/elrc_339} {"spanish-english website parallel corpus."}.

\bibitem[{Gehrmann et~al.(2021)Gehrmann, Adewumi, Aggarwal, Ammanamanchi, Anuoluwapo, Bosselut, Chandu, Clinciu, Das, Dhole et~al.}]{gehrmann2021gem}
Sebastian Gehrmann, Tosin Adewumi, Karmanya Aggarwal, Pawan~Sasanka Ammanamanchi, Aremu Anuoluwapo, Antoine Bosselut, Khyathi~Raghavi Chandu, Miruna Clinciu, Dipanjan Das, Kaustubh~D Dhole, et~al. 2021.
\newblock The gem benchmark: Natural language generation, its evaluation and metrics.
\newblock \emph{arXiv preprint arXiv:2102.01672}.

\bibitem[{Han et~al.(2022)Han, Schoelkopf, Zhao, Qi, Riddell, Benson, Sun, Zubova, Qiao, Burtell, Peng, Fan, Liu, Wong, Sailor, Ni, Nan, Kasai, Yu, Zhang, Joty, Fabbri, Kryscinski, Lin, Xiong, and Radev}]{nt_han2022folio}
Simeng Han, Hailey Schoelkopf, Yilun Zhao, Zhenting Qi, Martin Riddell, Luke Benson, Lucy Sun, Ekaterina Zubova, Yujie Qiao, Matthew Burtell, David Peng, Jonathan Fan, Yixin Liu, Brian Wong, Malcolm Sailor, Ansong Ni, Linyong Nan, Jungo Kasai, Tao Yu, Rui Zhang, Shafiq Joty, Alexander~R. Fabbri, Wojciech Kryscinski, Xi~Victoria Lin, Caiming Xiong, and Dragomir Radev. 2022.
\newblock \href {https://arxiv.org/abs/2209.00840} {Folio: Natural language reasoning with first-order logic}.
\newblock \emph{arXiv preprint arXiv:2209.00840}.

\bibitem[{Huang et~al.(2023{\natexlab{a}})Huang, Dong, Wang, Hao, Singhal, Ma, Lv, Cui, Mohammed, Liu et~al.}]{nt_huang2023language}
Shaohan Huang, Li~Dong, Wenhui Wang, Yaru Hao, Saksham Singhal, Shuming Ma, Tengchao Lv, Lei Cui, Owais~Khan Mohammed, Qiang Liu, et~al. 2023{\natexlab{a}}.
\newblock Language is not all you need: Aligning perception with language models.
\newblock \emph{arXiv preprint arXiv:2302.14045}.

\bibitem[{Huang et~al.(2023{\natexlab{b}})Huang, Bai, Zhu, Zhang, Zhang, Su, Liu, Lv, Zhang, Lei et~al.}]{huang2023ceval}
Yuzhen Huang, Yuzhuo Bai, Zhihao Zhu, Junlei Zhang, Jinghan Zhang, Tangjun Su, Junteng Liu, Chuancheng Lv, Yikai Zhang, Jiayi Lei, et~al. 2023{\natexlab{b}}.
\newblock C-eval: A multi-level multi-discipline chinese evaluation suite for foundation models.
\newblock \emph{arXiv preprint arXiv:2305.08322}.

\bibitem[{Islam et~al.(2022)Islam, Anik, and Islam}]{islam2022enhanced}
Md~Adnanul Islam, Md~Saidul~Hoque Anik, and ABM Alim~Al Islam. 2022.
\newblock An enhanced rbmt: When rbmt outperforms modern data-driven translators.
\newblock \emph{IETE Technical Review}, 39(6):1473--1484.

\bibitem[{Khashabi et~al.(2020)Khashabi, Min, Khot, Sabharwal, Tafjord, Clark, and Hajishirzi}]{khashabi2020unifiedqa}
Daniel Khashabi, Sewon Min, Tushar Khot, Ashish Sabharwal, Oyvind Tafjord, Peter Clark, and Hannaneh Hajishirzi. 2020.
\newblock Unifiedqa: Crossing format boundaries with a single qa system.
\newblock \emph{arXiv preprint arXiv:2005.00700}.

\bibitem[{Kwon et~al.(2023)Kwon, Li, Zhuang, Sheng, Zheng, Yu, Gonzalez, Zhang, and Stoica}]{kwon2023efficient}
Woosuk Kwon, Zhuohan Li, Siyuan Zhuang, Ying Sheng, Lianmin Zheng, Cody~Hao Yu, Joseph~E. Gonzalez, Hao Zhang, and Ion Stoica. 2023.
\newblock Efficient memory management for large language model serving with pagedattention.
\newblock In \emph{Proceedings of the ACM SIGOPS 29th Symposium on Operating Systems Principles}.

\bibitem[{Lake and Baroni(2017)}]{nt_lake2017generalization}
Brenden~M Lake and Marco Baroni. 2017.
\newblock Generalization without systematicity: On the compositional skills of sequence-to-sequence recurrent networks. arxiv.

\bibitem[{Li et~al.(2023)Li, Zhang, Koto, Yang, Zhao, Gong, Duan, and Baldwin}]{li2023cmmlu}
Haonan Li, Yixuan Zhang, Fajri Koto, Yifei Yang, Hai Zhao, Yeyun Gong, Nan Duan, and Timothy Baldwin. 2023.
\newblock Cmmlu: Measuring massive multitask language understanding in chinese.
\newblock \emph{arXiv preprint arXiv:2306.09212}.

\bibitem[{Liang et~al.(2023)Liang, Wu, Song, Wu, Xia, Liu, Ou, Lu, Ji, Mao et~al.}]{nt_liang2023taskmatrix}
Yaobo Liang, Chenfei Wu, Ting Song, Wenshan Wu, Yan Xia, Yu~Liu, Yang Ou, Shuai Lu, Lei Ji, Shaoguang Mao, et~al. 2023.
\newblock Taskmatrix. ai: Completing tasks by connecting foundation models with millions of apis.
\newblock \emph{arXiv preprint arXiv:2303.16434}.

\bibitem[{Liu et~al.(2023)Liu, Iter, Xu, Wang, Xu, and Zhu}]{liu2023gpteval}
Yang Liu, Dan Iter, Yichong Xu, Shuohang Wang, Ruochen Xu, and Chenguang Zhu. 2023.
\newblock Gpteval: Nlg evaluation using gpt-4 with better human alignment.
\newblock \emph{arXiv preprint arXiv:2303.16634}.

\bibitem[{Market(2018)}]{nt_shujujishi2018}
Data Market. 2018.
\newblock shujujishi.com.
\newblock \url{http://shujujishi.com/dataset/a037ab86-7727-487b-9a46-2936b0be076b.html}.
\newblock Accessed 16-02-2024.

\bibitem[{Mishra et~al.(2021)Mishra, Khashabi, Baral, and Hajishirzi}]{mishra2021cross}
Swaroop Mishra, Daniel Khashabi, Chitta Baral, and Hannaneh Hajishirzi. 2021.
\newblock Cross-task generalization via natural language crowdsourcing instructions.
\newblock \emph{arXiv preprint arXiv:2104.08773}.

\bibitem[{Oda(2016)}]{Oda2016}
Yusuke Oda. 2016.
\newblock Small parallel enja.
\newblock \url{https://github.com/odashi/small_parallel_enja}.

\bibitem[{Pei and Jurgens(2020)}]{nt_pei2020quantifying}
Jiaxin Pei and David Jurgens. 2020.
\newblock Quantifying intimacy in language.
\newblock \emph{arXiv preprint arXiv:2011.03020}.

\bibitem[{Pei and Jurgens(2021)}]{nt_pei2021measuring}
Jiaxin Pei and David Jurgens. 2021.
\newblock Measuring sentence-level and aspect-level (un)certainty in science communications.
\newblock In \emph{Proceedings of the 2021 Conference on Empirical Methods in Natural Language Processing (EMNLP)}.

\bibitem[{Perez-Almendros et~al.(2022)Perez-Almendros, Espinosa-Anke, and Schockaert}]{nt_perez-almendros-etal-2022-semeval}
Carla Perez-Almendros, Luis Espinosa-Anke, and Steven Schockaert. 2022.
\newblock \href {https://doi.org/10.18653/v1/2022.semeval-1.38} {{S}em{E}val-2022 task 4: Patronizing and condescending language detection}.
\newblock In \emph{Proceedings of the 16th International Workshop on Semantic Evaluation (SemEval-2022)}, pages 298--307, Seattle, United States. Association for Computational Linguistics.

\bibitem[{Raffel et~al.(2023)Raffel, Shazeer, Roberts, Lee, Narang, Matena, Zhou, Li, and Liu}]{raffel2023exploring}
Colin Raffel, Noam Shazeer, Adam Roberts, Katherine Lee, Sharan Narang, Michael Matena, Yanqi Zhou, Wei Li, and Peter~J. Liu. 2023.
\newblock \href {http://arxiv.org/abs/1910.10683} {Exploring the limits of transfer learning with a unified text-to-text transformer}.

\bibitem[{Rajpurkar et~al.(2018)Rajpurkar, Jia, and Liang}]{nt_rajpurkar2018know}
Pranav Rajpurkar, Robin Jia, and Percy Liang. 2018.
\newblock Know what you don't know: Unanswerable questions for squad.
\newblock \emph{arXiv preprint arXiv:1806.03822}.

\bibitem[{Ramasamy et~al.(2012)Ramasamy, Bojar, and {\v{Z}}abokrtsk{\'{y}}}]{biblio:RaBoMorphologicalProcessing2012}
Loganathan Ramasamy, Ond{\v{r}}ej Bojar, and Zden{\v{e}}k {\v{Z}}abokrtsk{\'{y}}. 2012.
\newblock Morphological processing for english-tamil statistical machine translation.
\newblock In \emph{Proceedings of the Workshop on Machine Translation and Parsing in Indian Languages ({MTPIL}-2012)}, pages 113--122.

\bibitem[{Reddy et~al.(2019)Reddy, Chen, and Manning}]{nt_reddy2019coqa}
Siva Reddy, Danqi Chen, and Christopher~D Manning. 2019.
\newblock Coqa: A conversational question answering challenge.
\newblock \emph{Transactions of the Association for Computational Linguistics}, 7:249--266.

\bibitem[{Sainz et~al.(2023)Sainz, Campos, Garc{\'\i}a-Ferrero, Etxaniz, de~Lacalle, and Agirre}]{sainz2023nlp_eval_trouble}
Oscar Sainz, Jon~Ander Campos, Iker Garc{\'\i}a-Ferrero, Julen Etxaniz, Oier~Lopez de~Lacalle, and Eneko Agirre. 2023.
\newblock Nlp evaluation in trouble: On the need to measure llm data contamination for each benchmark.
\newblock \emph{arXiv preprint arXiv:2310.18018}.

\bibitem[{Sanh et~al.(2021)Sanh, Webson, Raffel, Bach, Sutawika, Alyafeai, Chaffin, Stiegler, Scao, Raja et~al.}]{sanh2021T0}
Victor Sanh, Albert Webson, Colin Raffel, Stephen~H Bach, Lintang Sutawika, Zaid Alyafeai, Antoine Chaffin, Arnaud Stiegler, Teven~Le Scao, Arun Raja, et~al. 2021.
\newblock Multitask prompted training enables zero-shot task generalization.
\newblock \emph{arXiv preprint arXiv:2110.08207}.

\bibitem[{Sellam et~al.(2020)Sellam, Das, and Parikh}]{sellam2020bleurt}
Thibault Sellam, Dipanjan Das, and Ankur~P Parikh. 2020.
\newblock Bleurt: Learning robust metrics for text generation.
\newblock \emph{arXiv preprint arXiv:2004.04696}.

\bibitem[{Shah and Bakrola(2019)}]{shah2019neural}
Parth Shah and Vishvajit Bakrola. 2019.
\newblock Neural machine translation system of indic languages-an attention based approach.
\newblock In \emph{2019 Second International Conference on Advanced Computational and Communication Paradigms (ICACCP)}, pages 1--5. IEEE.

\bibitem[{Sun et~al.(2023)Sun, Zhang, He, Li, Cheng, Yan, Liu, Shao, Tang, Zhao, Chen, Zheng, Zhou, Li, Zhan, Zhou, Li, Yang, Wu, Yin, Huang, and Qiu}]{sun2023moss}
Tianxiang Sun, Xiaotian Zhang, Zhengfu He, Peng Li, Qinyuan Cheng, Hang Yan, Xiangyang Liu, Yunfan Shao, Qiong Tang, Xingjian Zhao, Ke~Chen, Yining Zheng, Zhejian Zhou, Ruixiao Li, Jun Zhan, Yunhua Zhou, Linyang Li, Xiaogui Yang, Lingling Wu, Zhangyue Yin, Xuanjing Huang, and Xipeng Qiu. 2023.
\newblock Moss: Training conversational language models from synthetic data.

\bibitem[{Talmor et~al.(2018)Talmor, Herzig, Lourie, and Berant}]{nt_talmor2018commonsenseqa}
Alon Talmor, Jonathan Herzig, Nicholas Lourie, and Jonathan Berant. 2018.
\newblock Commonsenseqa: A question answering challenge targeting commonsense knowledge.
\newblock \emph{arXiv preprint arXiv:1811.00937}.

\bibitem[{Tseng et~al.(2020)Tseng, Wu, Chang, Chen, and Hsu}]{nt_tseng2020development}
Yuen-Hsien Tseng, Wun-Syuan Wu, Chia-Yueh Chang, Hsueh-Chih Chen, and Wei-Lun Hsu. 2020.
\newblock Development and validation of a corpus for machine humor comprehension.
\newblock In \emph{Proceedings of the Twelfth Language Resources and Evaluation Conference}, pages 1346--1352.

\bibitem[{Wang et~al.(2020)Wang, Liang, Jin, Wang, Zhu, and Zhang}]{nt_wang-etal-2020-semeval}
Cunxiang Wang, Shuailong Liang, Yili Jin, Yilong Wang, Xiaodan Zhu, and Yue Zhang. 2020.
\newblock \href {https://doi.org/10.18653/v1/2020.semeval-1.39} {{S}em{E}val-2020 task 4: Commonsense validation and explanation}.
\newblock In \emph{Proceedings of the Fourteenth Workshop on Semantic Evaluation}, pages 307--321, Barcelona (online). International Committee for Computational Linguistics.

\bibitem[{Wang et~al.(2022{\natexlab{a}})Wang, Zhang, Zhang, Yang, Gao, Wu, Dong, He, Zhuo, Yang, Huang, Li, Wu, Lu, Zhu, Chen, Han, Pan, Wang, Wang, Wu, Zeng, Chen, Gan, and Zhang}]{fengshenbang}
Junjie Wang, Yuxiang Zhang, Lin Zhang, Ping Yang, Xinyu Gao, Ziwei Wu, Xiaoqun Dong, Junqing He, Jianheng Zhuo, Qi~Yang, Yongfeng Huang, Xiayu Li, Yanghan Wu, Junyu Lu, Xinyu Zhu, Weifeng Chen, Ting Han, Kunhao Pan, Rui Wang, Hao Wang, Xiaojun Wu, Zhongshen Zeng, Chongpei Chen, Ruyi Gan, and Jiaxing Zhang. 2022{\natexlab{a}}.
\newblock Fengshenbang 1.0: Being the foundation of chinese cognitive intelligence.
\newblock \emph{CoRR}, abs/2209.02970.

\bibitem[{Wang et~al.(2022{\natexlab{b}})Wang, Mishra, Alipoormolabashi, Kordi, Mirzaei, Arunkumar, Ashok, Dhanasekaran, Naik, Stap et~al.}]{wang2022super}
Yizhong Wang, Swaroop Mishra, Pegah Alipoormolabashi, Yeganeh Kordi, Amirreza Mirzaei, Anjana Arunkumar, Arjun Ashok, Arut~Selvan Dhanasekaran, Atharva Naik, David Stap, et~al. 2022{\natexlab{b}}.
\newblock Super-naturalinstructions: Generalization via declarative instructions on 1600+ nlp tasks.
\newblock \emph{arXiv preprint arXiv:2204.07705}.

\bibitem[{Wang et~al.(2024)Wang, Liu, Liu, Yao, Huang, He, Li, Li, Che, Zhang, Wang, Wang, Pu, Xu, Fang, Zhao, Zhang, Huang, Lu, Peng, Zheng, Wang, Yang, He, Jiang, Xie, Zhang, Li, Shi, Fu, Zhang, Huang, Xiong, Zhang, Wang, and Song}]{DBLP:journals/corr/abs-2401-03804}
Zihan Wang, Xinzhang Liu, Shixuan Liu, Yitong Yao, Yuyao Huang, Zhongjiang He, Xuelong Li, Yongxiang Li, Zhonghao Che, Zhaoxi Zhang, Yan Wang, Xin Wang, Luwen Pu, Huihan Xu, Ruiyu Fang, Yu~Zhao, Jie Zhang, Xiaomeng Huang, Zhilong Lu, Jiaxin Peng, Wenjun Zheng, Shiquan Wang, Bingkai Yang, Xuewei He, Zhuoru Jiang, Qiyi Xie, Yanhan Zhang, Zhongqiu Li, Lingling Shi, Weiwei Fu, Yin Zhang, Zilu Huang, Sishi Xiong, Yuxiang Zhang, Chao Wang, and Shuangyong Song. 2024.
\newblock Telechat technical report.
\newblock \emph{CoRR}, abs/2401.03804.

\bibitem[{Wei et~al.(2021)Wei, Bosma, Zhao, Guu, Yu, Lester, Du, Dai, and Le}]{wei2021FLAN}
Jason Wei, Maarten Bosma, Vincent~Y Zhao, Kelvin Guu, Adams~Wei Yu, Brian Lester, Nan Du, Andrew~M Dai, and Quoc~V Le. 2021.
\newblock Finetuned language models are zero-shot learners.
\newblock \emph{arXiv preprint arXiv:2109.01652}.

\bibitem[{Wikipedia(2024)}]{nt_wiki_cninternet}
Wikipedia. 2024.
\newblock {List of China Mainland Internet Language} --- {W}ikipedia{,} the free encyclopedia.
\newblock \url{http://zh.wikipedia.org/w/index.php?title=\%E4\%B8\%AD\%E5\%9B\%BD\%E5\%A4\%A7\%E9\%99\%86\%E7\%BD\%91\%E7\%BB\%9C\%E7\%94\%A8\%E8\%AF\%AD\%E5\%88\%97\%E8\%A1\%A8&oldid=81048845}.

\bibitem[{Xi et~al.(2022)Xi, Lv, Liu, Ye, Yang, and Wan}]{nt_xi2022musied}
Xiangyu Xi, Jianwei Lv, Shuaipeng Liu, Wei Ye, Fan Yang, and Guanglu Wan. 2022.
\newblock Musied: A benchmark for event detection from multi-source heterogeneous informal texts.
\newblock \emph{arXiv preprint arXiv:2211.13896}.

\bibitem[{Xu et~al.(2020)Xu, Hu, Zhang, Li, Cao, Li, Xu, Sun, Yu, Yu et~al.}]{xu2020clue}
Liang Xu, Hai Hu, Xuanwei Zhang, Lu~Li, Chenjie Cao, Yudong Li, Yechen Xu, Kai Sun, Dian Yu, Cong Yu, et~al. 2020.
\newblock Clue: A chinese language understanding evaluation benchmark.
\newblock \emph{arXiv preprint arXiv:2004.05986}.

\bibitem[{Xu et~al.(2021)Xu, Lu, Yuan, Zhang, Xu, Yuan, Wei, Pan, Tian, Qin et~al.}]{nt_xu2021fewclue}
Liang Xu, Xiaojing Lu, Chenyang Yuan, Xuanwei Zhang, Huilin Xu, Hu~Yuan, Guoao Wei, Xiang Pan, Xin Tian, Libo Qin, et~al. 2021.
\newblock Fewclue: A chinese few-shot learning evaluation benchmark.
\newblock \emph{arXiv preprint arXiv:2107.07498}.

\bibitem[{Yang et~al.(2018)Yang, Qi, Zhang, Bengio, Cohen, Salakhutdinov, and Manning}]{nt_yang2018hotpotqa}
Zhilin Yang, Peng Qi, Saizheng Zhang, Yoshua Bengio, William~W. Cohen, Ruslan Salakhutdinov, and Christopher~D. Manning. 2018.
\newblock {HotpotQA}: A dataset for diverse, explainable multi-hop question answering.
\newblock In \emph{Conference on Empirical Methods in Natural Language Processing ({EMNLP})}.

\bibitem[{Yao et~al.(2021)Yao, Dong, Guan, Cao, Zhang, Xiao, Wang, Qi, Bao, Nie et~al.}]{yao2021cuge}
Yuan Yao, Qingxiu Dong, Jian Guan, Boxi Cao, Zhengyan Zhang, Chaojun Xiao, Xiaozhi Wang, Fanchao Qi, Junwei Bao, Jinran Nie, et~al. 2021.
\newblock Cuge: A chinese language understanding and generation evaluation benchmark.
\newblock \emph{arXiv preprint arXiv:2112.13610}.

\bibitem[{Ye et~al.(2021)Ye, Lin, and Ren}]{ye2021crossfit}
Qinyuan Ye, Bill~Yuchen Lin, and Xiang Ren. 2021.
\newblock Crossfit: A few-shot learning challenge for cross-task generalization in nlp.
\newblock \emph{arXiv preprint arXiv:2104.08835}.

\bibitem[{Zellers et~al.(2019)Zellers, Holtzman, Bisk, Farhadi, and Choi}]{nt_zellers2019hellaswag}
Rowan Zellers, Ari Holtzman, Yonatan Bisk, Ali Farhadi, and Yejin Choi. 2019.
\newblock Hellaswag: Can a machine really finish your sentence?
\newblock In \emph{Proceedings of the 57th Annual Meeting of the Association for Computational Linguistics}.

\bibitem[{Zeng et~al.(2023)Zeng, Liu, Du, Wang, Lai, Ding, Yang, Xu, Zheng, Xia, Tam, Ma, Xue, Zhai, Chen, Liu, Zhang, Dong, and Tang}]{zeng2023glm-130b}
Aohan Zeng, Xiao Liu, Zhengxiao Du, Zihan Wang, Hanyu Lai, Ming Ding, Zhuoyi Yang, Yifan Xu, Wendi Zheng, Xiao Xia, Weng~Lam Tam, Zixuan Ma, Yufei Xue, Jidong Zhai, Wenguang Chen, Zhiyuan Liu, Peng Zhang, Yuxiao Dong, and Jie Tang. 2023.
\newblock \href {https://openreview.net/forum?id=-Aw0rrrPUF} {{GLM}-130b: An open bilingual pre-trained model}.
\newblock In \emph{The Eleventh International Conference on Learning Representations (ICLR)}.

\bibitem[{Zhang et~al.(2023{\natexlab{a}})Zhang, Li, Wu, Zhang, Lin, Geng, Wang, and Fu}]{nt_zhang2023corgi}
Ge~Zhang, Yizhi Li, Yaoyao Wu, Linyuan Zhang, Chenghua Lin, Jiayi Geng, Shi Wang, and Jie Fu. 2023{\natexlab{a}}.
\newblock Corgi-pm: A chinese corpus for gender bias probing and mitigation.
\newblock \emph{arXiv preprint arXiv:2301.00395}.

\bibitem[{Zhang et~al.(2023{\natexlab{b}})Zhang, Cahyawijaya, Cruz, and Aji}]{zhang2023multilingual}
Ruochen Zhang, Samuel Cahyawijaya, Jan Christian~Blaise Cruz, and Alham~Fikri Aji. 2023{\natexlab{b}}.
\newblock Multilingual large language models are not (yet) code-switchers.
\newblock \emph{arXiv preprint arXiv:2305.14235}.

\bibitem[{Zhao et~al.(2023)Zhao, Li, Hou, Zhao, Tian, Liu, Chen, Sun, Liu, Mao, Guo, Guo, Wu, Zhu, Shi, Chen, Huang, Chen, Liu, Li, Chen, Sun, Kang, Du, Shen, and Yan}]{DBLP:conf/acl/ZhaoLHZTLCSLMGG23}
Zhe Zhao, Yudong Li, Cheng Hou, Jing Zhao, Rong Tian, Weijie Liu, Yiren Chen, Ningyuan Sun, Haoyan Liu, Weiquan Mao, Han Guo, Weigang Guo, Taiqiang Wu, Tao Zhu, Wenhang Shi, Chen Chen, Shan Huang, Sihong Chen, Liqun Liu, Feifei Li, Xiaoshuai Chen, Xingwu Sun, Zhanhui Kang, Xiaoyong Du, Linlin Shen, and Kimmo Yan. 2023.
\newblock Tencentpretrain: {A} scalable and flexible toolkit for pre-training models of different modalities.
\newblock In \emph{{ACL} (demo)}, pages 217--225. Association for Computational Linguistics.

\bibitem[{Ziems et~al.(2022)Ziems, Li, Zhang, and Yang}]{nt_ziems-etal-2022-positive-frames}
Caleb Ziems, Minzhi Li, Anthony Zhang, and Diyi Yang. 2022.
\newblock Inducing positive perspectives with text reframing.
\newblock In \emph{Proceedings of the 60th Annual Meeting of the Association for Computational Linguistics}, Online and Dublin, Ireland. Association for Computational Linguistics.

\end{thebibliography}
\bibliographystyle{acl_natbib}

\appendix

\section{Task Details}

\subsection{Full List of Tasks and Evaluation}
\label{apdx:tasks}
We provide a full list of the task names and the source for input-output annotation in this subsection. The comprehensive task descriptions and the corresponding evaluation prompts can be found in the supplementary files.

\begin{table}[!htp]\centering
\caption{Full task list and source (1/3).}\label{tab: }
\scriptsize
\begin{tabular}{p{4cm}|p{2.5cm}}\toprule
\textbf{Task ID \& Name} &\textbf{Source} \\\midrule
0 Negotiation Strategy Detection &SNI~\cite{wang2022super} \\
1 Grammar Error Correction &SNI~\cite{wang2022super} \\
2 Overlap Extraction &SNI~\cite{wang2022super} \\
3 Commonsense &SNI~\cite{wang2022super} \\
4 Data to Text &SNI~\cite{wang2022super} \\
5 Keyword Tagging &SNI~\cite{wang2022super} \\
6 Answerability Classification &SNI~\cite{wang2022super} \\
7 Dialogue Act Recognition &SNI~\cite{wang2022super} \\
8 Cause Effect Classification &SNI~\cite{wang2022super} \\
9 Question Rewriting &SNI~\cite{wang2022super} \\
10 Textual Entailment &SNI~\cite{wang2022super} \\
11 Coreference Resolution &SNI~\cite{wang2022super} \\
12 Title Generation &SNI~\cite{wang2022super} \\
13 Entity Relation Classification &SNI~\cite{wang2022super} \\
14 Punctuation Error Detection &SNI~\cite{wang2022super} \\
15 Style Transfer &SNI~\cite{wang2022super} \\
16 Sentence Expansion &SNI~\cite{wang2022super} \\
17 Poem Generation &SNI~\cite{wang2022super} \\
18 Discourse Relation Classification &SNI~\cite{wang2022super} \\
19 Mathematics &SNI~\cite{wang2022super} \\
20 Text Simplification &SNI~\cite{wang2022super} \\
21 Sentence Compression &SNI~\cite{wang2022super} \\
22 Spelling Error Detection &SNI~\cite{wang2022super} \\
23 Irony Detection &SNI~\cite{wang2022super} \\
24 Number Conversion &SNI~\cite{wang2022super} \\
25 Word Relation Classification &SNI~\cite{wang2022super} \\
26 Paraphrasing &SNI~\cite{wang2022super} \\
27 Grammar Error Detection &SNI~\cite{wang2022super} \\
28 Text Matching &SNI~\cite{wang2022super} \\
29 Fill in The Blank &SNI~\cite{wang2022super} \\
30 Speaker Relation Classification &SNI~\cite{wang2022super} \\
31 Entity Generation &SNI~\cite{wang2022super} \\
32 Summarization &SNI~\cite{wang2022super} \\
33 Spam Classification &SNI~\cite{wang2022super} \\
34 Stereotype Detection &SNI~\cite{wang2022super} \\
35 Dialogue State Tracking &SNI~\cite{wang2022super} \\
36 Dialogue State Tracking &SNI~\cite{wang2022super} \\
37 Sentence Perturbation &SNI~\cite{wang2022super} \\
38 Text Quality Evaluation &SNI~\cite{wang2022super} \\
39 Linguistic Probing &SNI~\cite{wang2022super} \\
40 Information Extraction &SNI~\cite{wang2022super} \\
41 Emotion Prediction &SNI~\cite{wang2022super} \\
42 Discourse Connective Identification &SNI~\cite{wang2022super} \\
43 Question Generation &SNI~\cite{wang2022super} \\
44 Stance Detection &SNI~\cite{wang2022super} \\
45 Sentiment Analysis &SNI~\cite{wang2022super} \\
46 Story Composition &SNI~\cite{wang2022super} \\
47 Program Execution &SNI~\cite{wang2022super} \\
48 Gender Classification &SNI~\cite{wang2022super} \\
49 Named Entity Recognition &SNI~\cite{wang2022super} \\
50 Toxic Language Detection &SNI~\cite{wang2022super} \\
51 Question Decomposition &SNI~\cite{wang2022super} \\
52 Sentence Ordering &SNI~\cite{wang2022super} \\
53 Text to Code &SNI~\cite{wang2022super} \\
54 Fact Verification &SNI~\cite{wang2022super} \\
55 Speaker Identification &SNI~\cite{wang2022super} \\
56 Answer Verification &SNI~\cite{wang2022super} \\
57 Wrong Candidate Generation &SNI~\cite{wang2022super} \\
58 Dialogue Generation &SNI~\cite{wang2022super} \\
59 Text Completion &SNI~\cite{wang2022super} \\
60 Pos Tagging &SNI~\cite{wang2022super} \\
\bottomrule
\end{tabular}
\end{table}

\begin{table}[!htp]\centering
\caption{Full task list and source (2/3).}\label{tab: 2}
\scriptsize
\begin{tabular}{p{4cm}|p{2.5cm}}\toprule
\textbf{Task ID \& Name} &\textbf{Source} \\\midrule
61 Explanation &SNI~\cite{wang2022super} \\
62 Sentence Composition &SNI~\cite{wang2022super} \\
63 Question Understanding &SNI~\cite{wang2022super} \\
64 Intent Identification &SNI~\cite{wang2022super} \\
65 Word Semantics &SNI~\cite{wang2022super} \\
66 Code to Text &SNI~\cite{wang2022super} \\
67 Preposition Prediction &SNI~\cite{wang2022super} \\
68 Text Categorization &SNI~\cite{wang2022super} \\
69 Question Answering &SNI~\cite{wang2022super} \\
70 Commonsense Classification &N/A \\
71 Ancient Chinese Poem Retrieval &N/A \\
72 Ancient Chinese Translation &N/A \\
73 Chinese Rhyme Detection &N/A \\
74 Nationality Detection &N/A \\
75 Region Detection &N/A \\
76 Chinese Idiom Explanation &N/A \\
77 Name Allusion Detection &N/A \\
78 Chinese Ambiguity Sentence Location &N/A \\
79 Chinese Winograd Schema Challenge &FewCLUE~\cite{nt_xu2021fewclue} \\

80 Chinese Modern Abbreviation Explanation &Wikipedia~\cite{nt_wiki_cninternet} \\
81 Chinese Epigraph Detection &N/A \\
82 Chinese Dialect Translation &N/A \\
83 Chinese Attractions List &N/A \\
85 Chinese Typo Categorization &N/A \\
86 Chinese Fiction Characteristic Detection &N/A \\
87 Chinese Figurative Detection &N/A \\
88 Chinese Metaphor Explanation &N/A \\
89 Chinese Medicine Detection &N/A \\
90 Chinese Pinyin Detection &N/A \\
91 Chinese Wubi Written &N/A \\
92 Intimacy Score Prediction &\citet{nt_pei2020quantifying} \\
93 Sentence Level Uncertainty Judgement &\citet{nt_pei2021measuring} \\
94 Chinese Relative Identification &N/A \\
96 Chinese Heteronomous Language Detection &N/A \\
97 Code Debug &\url{https://blog.csdn.net} \\
98 Code Translate &\url{https://leetcode.cn} \\
99 Function Explanation &\url{https://www.liaoxuefeng.com/} \\
100 Bias Detoxication &CORGI-PM~\cite{nt_zhang2023corgi} \\
101 MultiLabel Chinese Humor Categorization &\citet{nt_tseng2020development} \\
102 Legal Term Retrieval &N/A \\
103 Patronizing Condescending Multilabel &\citet{nt_perez-almendros-etal-2022-semeval} \\
104 CommonSense Explanation &\citet{nt_wang-etal-2020-semeval} \\
105 Event Type Detection &MUSIED~\citep{nt_xi2022musied} \\
106 Argument Mining &N/A \\
107 Theory of Mind &Big-Bench Theory of Mind~\citep{nt_srivastava2023beyond} \\
108 Game Playing &Big-Bench Language Games~\citep{nt_srivastava2023beyond} \\
109 IQ Test &\citet{nt_huang2023language} \\
110 Joke Explanation &N/A \\
111 Role Playing &TRPG \url{https://bilibili.com}~\cite{nt_cote18textworld} \\
112 Text De-Identification &N/A \\
113 Outline Generation &N/A \\
114 Pros Cons Listing &N/A \\
115 Joke Telling &N/A \\
116 Affordance &N/A \\
\bottomrule
\end{tabular}
\end{table}

\begin{table}[!htp]\centering
\caption{Full task list and source  (2/3).}\label{tab: 2}
\scriptsize
\begin{tabular}{p{4cm}|p{2.5cm}}\toprule
\textbf{Task ID \& Name} &\textbf{Source} \\\midrule
117 Material Synthesis &\citet{nt_bara-etal-2021-mindcraft} \\
118 Tool use &Taskmatrix~\cite{nt_liang2023taskmatrix} \\
119 Concept Abstraction &N/A \\
120 Rhyme Aligned Generation &N/A \\
121 Advertising &N/A \\
122 Mind Tree Generation &N/A \\
123 First Order Logic &FOLIO~\cite{nt_han2022folio} \\
124 Critical Thinking &N/A \\
125 Empathy Detection &N/A \\
126 Social Norms Detection &Moral Stories~\cite{nt_emelin-etal-2021-moral} \\
127 Make Positive &\citet{nt_ziems-etal-2022-positive-frames} \\
128 Translate to Ancient Chinese &N/A \\
129 Recipe Generation &\citet{nt_shujujishi2018} \\
130 Imagination &N/A \\
131 Compositional Reasoning &\citet{nt_lake2017generalization} \\
132 Personality Detection &N/A \\
133 Table Generation &N/A \\
134 Flowchart Generation &N/A \\
135 Review Generation &N/A \\
136 Draw Figure with symbol &N/A \\
137 CommonsenseQA &CommonsenseQA~\cite{nt_talmor2018commonsenseqa} \\
138 ReadingComprehensionQA & \citet{nt_rajpurkar2018know} \\
139 DiscreteOperationQA & DROP\cite{nt_dua2019drop} \\
140 MultiHopQA & HotpotQA~\cite{nt_yang2018hotpotqa} \\
141 CommonsenseNLI & HellaSwag~\cite{nt_zellers2019hellaswag} \\
142 ConversationalQA & CoQA~\cite{nt_reddy2019coqa} \\
143 MathQA & GSM8K~\cite{cobbe2021gsm8k} \\
144 English translation & N/A \\
145 French translation & DiaBLa~\cite{bawden_DiaBLa-A-Corpus-of_2021} \\
146 Arabic translation &\citet{altammami-etal-2020-constructing} \\
147 Japanese translation &\citet{Oda2016} \\
148 Spanish translation &\citet{spanishen2017} \\
149 Bengali translation &\citet{islam2022enhanced} \\
150 Tamil translation &\citet{biblio:RaBoMorphologicalProcessing2012} \\
151 Gujarati translation &\citet{shah2019neural} \\
\bottomrule
\end{tabular}
\end{table}

\subsection{Category Description}\label{apdx:categories}
We provide the task category description in this subsection.

\paragraph{Chinese Culture (18).} Focuses on aspects unique to Chinese history, society, and language, therefore testing the model's understanding of cultural nuances.

\paragraph{Classification (21).}
Addresses classification tasks, such as determining correctness or whether something belongs to a specific category.

\paragraph{Code (5).} Tests the model's proficiency in understanding and generating computer code across various programming languages.

\paragraph{Commonsense (36).} Evaluates the model's grasp of general knowledge and everyday reasoning that humans consider obvious.

\paragraph{Creative Natural Language Generation (NLG) (21).} Measures the model's ability to produce imaginative and novel text outputs, ranging from stories to creative descriptions.

\paragraph{Evaluation (5).} Focuses on assessing other models or systems, therefore testing the ability to judge and provide feedback on performance.

\paragraph{Grammar (10).} Assesses the model's understanding of linguistic rules and its ability to apply them correctly in text generation.

\paragraph{Linguistic (24).} Involves tasks that test the model's understanding of language structure, including syntax, semantics, and morphology.

\paragraph{Motion Detection (6).} Uncommon in LLMs, this refers to tasks related to interpreting descriptions of motion or predicting outcomes based on textual motion descriptions.

\paragraph{Named Entity Recognition (NER) (12).} Involves identifying and categorizing key information (e.g., names, places, dates) within the text.

\paragraph{Natural Language Inference (NLI) (30).} Tests the model's ability to understand relationships between sentences, such as contradiction, entailment, and neutrality.

\paragraph{Question Answering (QA) (19).} Evaluates the model's ability to understand and respond to questions with accurate and relevant answers.

\paragraph{Reasoning (29).} Involves tasks that require logical thinking, problem-solving, and deduction to arrive at correct conclusions.

\paragraph{Role Playing (2).} Tests the model's ability to adopt personas or roles in conversational contexts, assessing its versatility in generating context-appropriate responses.

\paragraph{Sentiment (8).} Evaluates the model's ability to detect and interpret emotional tones in text, such as positive, negative, or neutral sentiments.

\paragraph{Structured Data (16).} Involves interpreting and generating responses based on structured information such as tables, charts, and databases.

\paragraph{Style Transfer (20).} Tests the model's ability to convert text from one stylistic or tonal form to another while retaining the original content's meaning.

\paragraph{Summarization (9).} Assesses the model's ability to condense longer texts into shorter, coherent summaries capturing the essential points.

\paragraph{Toxicity (3).} Focuses on identifying and mitigating harmful or stereotypical content in text generation.

\paragraph{Translation (9).} Evaluates the model's ability to accurately translate text between languages, testing its linguistic versatility and understanding.

\section{CIF-Bench Results in Public Split}\label{apdx:cif_public_scores}
We provide the category-based results in \textit{public} split here in \autoref{tab: overall result 1 public}.

\begin{table*}[!h]\centering
\caption{Overall Results in CIF-Bench \textit{Public} Split with Single Instruction. The first column is the average score across \textit{all} the tasks, and the rest columns are average scores grouped by task categories. The cells are highlighted with fading colors from \colorbox[HTML]{fbfada}{maximum} to \colorbox[HTML]{b4b4b8}{minimum} in a column.}\label{tab: overall result 1 public}
\scriptsize
\scalebox{0.8}{
\begin{tabular}{lrrrrrrrrrrrr}\toprule
\multicolumn{1}{l}{\textbf{Model Name}} &\multicolumn{1}{p{0.8cm}}{\textbf{Overall}} &\multicolumn{1}{p{1.0cm}}{\textbf{Chinese Culture}} &\multicolumn{1}{p{1.2cm}}{\textbf{Classification}} &\multicolumn{1}{p{0.8cm}}{\textbf{Code}} &\multicolumn{1}{p{1.2cm}}{\textbf{Commonsense}} &\multicolumn{1}{p{1.1cm}}{\textbf{Creative NLG}} &\multicolumn{1}{p{1.0cm}}{\textbf{Evaluation}} &\multicolumn{1}{p{1.0cm}}{\textbf{Grammar}} &\multicolumn{1}{p{1.0cm}}{\textbf{Linguistic}} &\multicolumn{1}{p{1.0cm}}{\textbf{Motion Detection}} &\multicolumn{1}{p{0.8cm}}{\textbf{NER}} \\\midrule
Qwen-72B-Chat &\cellcolor[HTML]{fbfada}0.589 &\cellcolor[HTML]{fbfada}0.512 &\cellcolor[HTML]{fbfada}0.716 &\cellcolor[HTML]{e6e5d9}0.444 &\cellcolor[HTML]{fbfada}0.706 &\cellcolor[HTML]{fbfada}0.587 &\cellcolor[HTML]{f1f0d9}0.661 &\cellcolor[HTML]{fbfada}0.424 &\cellcolor[HTML]{fbfada}0.521 &\cellcolor[HTML]{fbfada}0.694 &\cellcolor[HTML]{fbfada}0.515 \\
Qwen-14B-Chat &\cellcolor[HTML]{f6f4d9}0.564 &\cellcolor[HTML]{f7f6d9}0.481 &\cellcolor[HTML]{f5f4d9}0.678 &\cellcolor[HTML]{dfddd6}0.416 &\cellcolor[HTML]{f4f3d9}0.657 &\cellcolor[HTML]{f3f2d9}0.567 &\cellcolor[HTML]{f3f2d9}0.669 &\cellcolor[HTML]{f4f3d9}0.396 &\cellcolor[HTML]{f4f3d9}0.485 &\cellcolor[HTML]{f4f3d9}0.663 &\cellcolor[HTML]{f4f3d9}0.486 \\
Deepseek-LLM-67B-Chat &\cellcolor[HTML]{eeedd9}0.526 &\cellcolor[HTML]{f6f5d9}0.477 &\cellcolor[HTML]{ecebd9}0.617 &\cellcolor[HTML]{d7d5d0}0.364 &\cellcolor[HTML]{eeecd9}0.609 &\cellcolor[HTML]{f0eed9}0.559 &\cellcolor[HTML]{e1dfd7}0.573 &\cellcolor[HTML]{f0eed9}0.374 &\cellcolor[HTML]{f0eed9}0.458 &\cellcolor[HTML]{edecd9}0.631 &\cellcolor[HTML]{f6f4d9}0.493 \\
GPT-3.5-Public-SFT &\cellcolor[HTML]{edecd9}0.522 &\cellcolor[HTML]{e3e1d9}0.316 &\cellcolor[HTML]{ecead9}0.611 &\cellcolor[HTML]{fbfada}0.492 &\cellcolor[HTML]{eae8d9}0.578 &\cellcolor[HTML]{e8e6d9}0.538 &\cellcolor[HTML]{edecd9}0.639 &\cellcolor[HTML]{f0efd9}0.377 &\cellcolor[HTML]{eeecd9}0.447 &\cellcolor[HTML]{e2e0d8}0.580 &\cellcolor[HTML]{f5f4d9}0.492 \\
Yi-34B-Chat &\cellcolor[HTML]{ecebd9}0.516 &\cellcolor[HTML]{f3f2d9}0.452 &\cellcolor[HTML]{ebe9d9}0.607 &\cellcolor[HTML]{e4e2d9}0.437 &\cellcolor[HTML]{f0eed9}0.624 &\cellcolor[HTML]{e1dfd7}0.516 &\cellcolor[HTML]{dcdad4}0.545 &\cellcolor[HTML]{d6d4d0}0.254 &\cellcolor[HTML]{e2e0d8}0.382 &\cellcolor[HTML]{f6f4d9}0.671 &\cellcolor[HTML]{e0ded7}0.398 \\
Baichuan2-13B-Chat &\cellcolor[HTML]{ebead9}0.512 &\cellcolor[HTML]{f2f1d9}0.446 &\cellcolor[HTML]{edecd9}0.623 &\cellcolor[HTML]{dddcd5}0.403 &\cellcolor[HTML]{edebd9}0.600 &\cellcolor[HTML]{dedcd5}0.505 &\cellcolor[HTML]{e2e0d8}0.582 &\cellcolor[HTML]{ebe9d9}0.352 &\cellcolor[HTML]{e9e8d9}0.423 &\cellcolor[HTML]{eeecd9}0.633 &\cellcolor[HTML]{e9e7d9}0.435 \\
Tigerbot-13B-Chat &\cellcolor[HTML]{e8e6d9}0.494 &\cellcolor[HTML]{e7e5d9}0.350 &\cellcolor[HTML]{e4e2d9}0.558 &\cellcolor[HTML]{e8e6d9}0.447 &\cellcolor[HTML]{edebd9}0.599 &\cellcolor[HTML]{e4e2d9}0.528 &\cellcolor[HTML]{faf9d9}0.707 &\cellcolor[HTML]{ebe9d9}0.352 &\cellcolor[HTML]{eeecd9}0.447 &\cellcolor[HTML]{dcdbd4}0.551 &\cellcolor[HTML]{f7f6d9}0.498 \\
Chinese-Alpaca-2-13B &\cellcolor[HTML]{e7e6d9}0.492 &\cellcolor[HTML]{d1cfcc}0.260 &\cellcolor[HTML]{e6e4d9}0.572 &\cellcolor[HTML]{e2e0d8}0.434 &\cellcolor[HTML]{e4e2d9}0.533 &\cellcolor[HTML]{f1f0d9}0.562 &\cellcolor[HTML]{e1dfd7}0.574 &\cellcolor[HTML]{e3e1d9}0.318 &\cellcolor[HTML]{e8e7d9}0.417 &\cellcolor[HTML]{ecead9}0.624 &\cellcolor[HTML]{f0eed9}0.467 \\
Chinese-Alpaca-33B &\cellcolor[HTML]{e6e4d9}0.484 &\cellcolor[HTML]{d5d4cf}0.274 &\cellcolor[HTML]{e2e0d8}0.546 &\cellcolor[HTML]{f1f0d9}0.470 &\cellcolor[HTML]{e3e1d9}0.527 &\cellcolor[HTML]{e9e7d9}0.540 &\cellcolor[HTML]{faf8d9}0.703 &\cellcolor[HTML]{e6e5d9}0.332 &\cellcolor[HTML]{e2e0d8}0.382 &\cellcolor[HTML]{e3e1d9}0.582 &\cellcolor[HTML]{efeed9}0.464 \\
Ziya-Llama-13B &\cellcolor[HTML]{e5e3d9}0.479 &\cellcolor[HTML]{dad8d2}0.287 &\cellcolor[HTML]{e3e1d9}0.550 &\cellcolor[HTML]{e0dfd7}0.422 &\cellcolor[HTML]{e3e1d9}0.523 &\cellcolor[HTML]{edebd9}0.551 &\cellcolor[HTML]{efeed9}0.650 &\cellcolor[HTML]{dedcd6}0.294 &\cellcolor[HTML]{e3e1d9}0.384 &\cellcolor[HTML]{e9e7d9}0.610 &\cellcolor[HTML]{e9e7d9}0.437 \\
Chinese-Llama2-Linly-13B &\cellcolor[HTML]{e5e3d9}0.479 &\cellcolor[HTML]{d9d8d2}0.286 &\cellcolor[HTML]{edecd9}0.623 &\cellcolor[HTML]{e5e3d9}0.439 &\cellcolor[HTML]{e6e4d9}0.549 &\cellcolor[HTML]{e7e5d9}0.535 &\cellcolor[HTML]{ebe9d9}0.626 &\cellcolor[HTML]{dddbd4}0.286 &\cellcolor[HTML]{e6e4d9}0.403 &\cellcolor[HTML]{e4e2d9}0.587 &\cellcolor[HTML]{f0efd9}0.468 \\
Tigerbot-7B-Chat &\cellcolor[HTML]{e5e3d9}0.478 &\cellcolor[HTML]{e7e6d9}0.354 &\cellcolor[HTML]{dfddd6}0.528 &\cellcolor[HTML]{e5e3d9}0.440 &\cellcolor[HTML]{e9e7d9}0.570 &\cellcolor[HTML]{e9e7d9}0.540 &\cellcolor[HTML]{fbfada}0.708 &\cellcolor[HTML]{e3e1d9}0.314 &\cellcolor[HTML]{ebe9d9}0.430 &\cellcolor[HTML]{d8d6d1}0.528 &\cellcolor[HTML]{e4e2d9}0.413 \\
ChatGLM3-6B &\cellcolor[HTML]{e3e1d9}0.472 &\cellcolor[HTML]{e3e1d9}0.321 &\cellcolor[HTML]{d8d7d1}0.488 &\cellcolor[HTML]{e3e1d9}0.436 &\cellcolor[HTML]{e3e1d9}0.527 &\cellcolor[HTML]{dddcd5}0.503 &\cellcolor[HTML]{e3e1d9}0.588 &\cellcolor[HTML]{dddcd5}0.290 &\cellcolor[HTML]{d4d3cf}0.328 &\cellcolor[HTML]{e1dfd7}0.574 &\cellcolor[HTML]{e4e2d9}0.415 \\
Chinese-Alpaca-13B &\cellcolor[HTML]{e3e1d9}0.471 &\cellcolor[HTML]{d2d1cd}0.264 &\cellcolor[HTML]{e3e1d9}0.553 &\cellcolor[HTML]{e6e4d9}0.443 &\cellcolor[HTML]{d9d8d2}0.495 &\cellcolor[HTML]{e3e1d9}0.525 &\cellcolor[HTML]{e3e1d9}0.587 &\cellcolor[HTML]{e7e5d9}0.334 &\cellcolor[HTML]{e4e3d9}0.394 &\cellcolor[HTML]{f2f0d9}0.653 &\cellcolor[HTML]{edecd9}0.457 \\
ChatGLM2-6B &\cellcolor[HTML]{e1dfd8}0.464 &\cellcolor[HTML]{e5e3d9}0.334 &\cellcolor[HTML]{e0ded7}0.532 &\cellcolor[HTML]{e3e1d9}0.436 &\cellcolor[HTML]{e3e1d9}0.522 &\cellcolor[HTML]{e4e2d9}0.527 &\cellcolor[HTML]{efeed9}0.651 &\cellcolor[HTML]{e3e1d9}0.314 &\cellcolor[HTML]{e5e3d9}0.395 &\cellcolor[HTML]{d9d8d2}0.536 &\cellcolor[HTML]{e1dfd7}0.402 \\
Chinese-Alpaca-7B &\cellcolor[HTML]{dddbd4}0.452 &\cellcolor[HTML]{c9c8c7}0.237 &\cellcolor[HTML]{e1dfd7}0.536 &\cellcolor[HTML]{e4e2d9}0.438 &\cellcolor[HTML]{d5d4cf}0.484 &\cellcolor[HTML]{dddcd5}0.502 &\cellcolor[HTML]{f4f2d9}0.672 &\cellcolor[HTML]{e3e1d9}0.318 &\cellcolor[HTML]{e4e2d9}0.389 &\cellcolor[HTML]{f2f0d9}0.652 &\cellcolor[HTML]{dfddd6}0.394 \\
Chinese-Alpaca-2-7B &\cellcolor[HTML]{dbdad4}0.448 &\cellcolor[HTML]{cecdca}0.251 &\cellcolor[HTML]{d6d4d0}0.472 &\cellcolor[HTML]{e3e1d9}0.435 &\cellcolor[HTML]{d4d3cf}0.480 &\cellcolor[HTML]{e6e4d9}0.532 &\cellcolor[HTML]{e1dfd8}0.577 &\cellcolor[HTML]{d9d7d2}0.268 &\cellcolor[HTML]{dad8d2}0.348 &\cellcolor[HTML]{e6e4d9}0.596 &\cellcolor[HTML]{e8e6d9}0.431 \\
Chinese-Llama2-Linly-7B &\cellcolor[HTML]{d9d8d2}0.443 &\cellcolor[HTML]{d2d1cd}0.264 &\cellcolor[HTML]{e4e2d9}0.558 &\cellcolor[HTML]{e0ded7}0.419 &\cellcolor[HTML]{dad9d3}0.497 &\cellcolor[HTML]{e2e0d8}0.522 &\cellcolor[HTML]{f2f1d9}0.664 &\cellcolor[HTML]{d2d1cd}0.236 &\cellcolor[HTML]{e2e0d8}0.381 &\cellcolor[HTML]{e5e3d9}0.593 &\cellcolor[HTML]{dbdad3}0.381 \\
Qwen-7B-Chat &\cellcolor[HTML]{d9d7d2}0.442 &\cellcolor[HTML]{e2e0d8}0.313 &\cellcolor[HTML]{e3e1d9}0.549 &\cellcolor[HTML]{dddcd5}0.404 &\cellcolor[HTML]{e2e0d8}0.520 &\cellcolor[HTML]{e1dfd7}0.515 &\cellcolor[HTML]{eeedd9}0.646 &\cellcolor[HTML]{d4d2ce}0.244 &\cellcolor[HTML]{e7e6d9}0.411 &\cellcolor[HTML]{e0ded7}0.570 &\cellcolor[HTML]{d8d6d1}0.368 \\
ChatGLM-6B &\cellcolor[HTML]{d8d7d1}0.440 &\cellcolor[HTML]{e2e0d8}0.311 &\cellcolor[HTML]{dad9d3}0.499 &\cellcolor[HTML]{e7e5d9}0.446 &\cellcolor[HTML]{d5d4cf}0.484 &\cellcolor[HTML]{ecead9}0.548 &\cellcolor[HTML]{dedcd5}0.558 &\cellcolor[HTML]{dbd9d3}0.278 &\cellcolor[HTML]{e2e0d8}0.382 &\cellcolor[HTML]{cfcecb}0.484 &\cellcolor[HTML]{dddbd4}0.386 \\
Baichuan-13B-Chat &\cellcolor[HTML]{d3d2ce}0.426 &\cellcolor[HTML]{e7e6d9}0.355 &\cellcolor[HTML]{cccbc9}0.416 &\cellcolor[HTML]{d6d5d0}0.361 &\cellcolor[HTML]{e1dfd7}0.516 &\cellcolor[HTML]{c9c8c6}0.416 &\cellcolor[HTML]{dfddd6}0.564 &\cellcolor[HTML]{e5e3d9}0.324 &\cellcolor[HTML]{e0ded7}0.374 &\cellcolor[HTML]{bababc}0.380 &\cellcolor[HTML]{dfddd6}0.394 \\
Yi-6B-Chat &\cellcolor[HTML]{d1d0cc}0.420 &\cellcolor[HTML]{e3e1d9}0.320 &\cellcolor[HTML]{d0cfcc}0.439 &\cellcolor[HTML]{dcdad4}0.395 &\cellcolor[HTML]{d8d6d1}0.489 &\cellcolor[HTML]{d1d0cc}0.449 &\cellcolor[HTML]{d3d2ce}0.493 &\cellcolor[HTML]{d1d0cc}0.230 &\cellcolor[HTML]{cbcac8}0.293 &\cellcolor[HTML]{e4e2d9}0.587 &\cellcolor[HTML]{d1cfcc}0.341 \\
CPM-Bee-10B &\cellcolor[HTML]{cfcecb}0.415 &\cellcolor[HTML]{ebe9d9}0.382 &\cellcolor[HTML]{d3d1cd}0.455 &\cellcolor[HTML]{c9c8c7}0.284 &\cellcolor[HTML]{c3c3c3}0.431 &\cellcolor[HTML]{dfddd6}0.508 &\cellcolor[HTML]{b4b4b8}0.300 &\cellcolor[HTML]{e3e1d9}0.317 &\cellcolor[HTML]{deddd6}0.367 &\cellcolor[HTML]{d1d0cc}0.494 &\cellcolor[HTML]{e0ded6}0.397 \\
Moss-Moon-003-SFT &\cellcolor[HTML]{c9c8c7}0.399 &\cellcolor[HTML]{c8c7c6}0.233 &\cellcolor[HTML]{d4d3cf}0.465 &\cellcolor[HTML]{dbd9d3}0.389 &\cellcolor[HTML]{c2c1c2}0.427 &\cellcolor[HTML]{d9d7d2}0.482 &\cellcolor[HTML]{d6d5d0}0.509 &\cellcolor[HTML]{dad8d3}0.274 &\cellcolor[HTML]{dfddd6}0.369 &\cellcolor[HTML]{d7d6d1}0.526 &\cellcolor[HTML]{dcdbd4}0.385 \\
Belle-SFT-Public &\cellcolor[HTML]{c9c8c7}0.397 &\cellcolor[HTML]{bcbbbd}0.196 &\cellcolor[HTML]{dbd9d3}0.503 &\cellcolor[HTML]{d9d7d2}0.376 &\cellcolor[HTML]{c1c1c1}0.426 &\cellcolor[HTML]{d6d5d0}0.472 &\cellcolor[HTML]{dcdad4}0.543 &\cellcolor[HTML]{d9d7d2}0.269 &\cellcolor[HTML]{dfddd6}0.371 &\cellcolor[HTML]{d5d3cf}0.512 &\cellcolor[HTML]{d4d3cf}0.356 \\
Telechat-7B &\cellcolor[HTML]{b8b7ba}0.350 &\cellcolor[HTML]{b4b4b8}0.172 &\cellcolor[HTML]{b8b8bb}0.299 &\cellcolor[HTML]{e4e2d9}0.438 &\cellcolor[HTML]{b4b4b8}0.386 &\cellcolor[HTML]{d2d1cd}0.456 &\cellcolor[HTML]{c4c3c3}0.400 &\cellcolor[HTML]{bebdbf}0.138 &\cellcolor[HTML]{b4b4b8}0.202 &\cellcolor[HTML]{c0c0c1}0.412 &\cellcolor[HTML]{cbcac8}0.322 \\
Aquilachat-7B &\cellcolor[HTML]{b7b7ba}0.350 &\cellcolor[HTML]{bebdbf}0.203 &\cellcolor[HTML]{b4b4b8}0.270 &\cellcolor[HTML]{d5d4cf}0.357 &\cellcolor[HTML]{bab9bc}0.404 &\cellcolor[HTML]{d1cfcc}0.449 &\cellcolor[HTML]{c3c2c2}0.394 &\cellcolor[HTML]{b4b4b8}0.090 &\cellcolor[HTML]{c2c2c2}0.260 &\cellcolor[HTML]{b4b4b8}0.348 &\cellcolor[HTML]{cbcac8}0.322 \\
Baichuan2-7B-Chat &\cellcolor[HTML]{b4b4b8}0.339 &\cellcolor[HTML]{e6e4d9}0.345 &\cellcolor[HTML]{e9e8d9}0.595 &\cellcolor[HTML]{b4b4b8}0.154 &\cellcolor[HTML]{cbcac8}0.455 &\cellcolor[HTML]{b4b4b8}0.327 &\cellcolor[HTML]{d8d7d1}0.523 &\cellcolor[HTML]{edebd9}0.362 &\cellcolor[HTML]{dbd9d3}0.354 &\cellcolor[HTML]{cbcac8}0.466 &\cellcolor[HTML]{b4b4b8}0.233 \\
\midrule
\midrule
\textbf{Model Name} &\textbf{Overall} &\multicolumn{1}{p{0.6cm}}{\textbf{NLI}} &\multicolumn{1}{p{0.6cm}}{\textbf{QA}} &\multicolumn{1}{p{1.0cm}}{\textbf{Reasoning}} &\multicolumn{1}{p{1.0cm}}{\textbf{Role Playing}} &\multicolumn{1}{p{1.0cm}}{\textbf{Sentiment}} &\multicolumn{1}{p{1.0cm}}{\textbf{Structured Data}} &\multicolumn{1}{p{1.0cm}}{\textbf{Style Transfer}} &\multicolumn{1}{p{1.4cm}}{\textbf{Summarization}} & \multicolumn{1}{p{0.8cm}}{\textbf{Toxic}} &\multicolumn{1}{p{1.2cm}}{\textbf{Translation}} \\\midrule
Qwen-72B-Chat &\cellcolor[HTML]{fbfada}0.589 &\cellcolor[HTML]{fbfada}0.695 &\cellcolor[HTML]{fbfada}0.668 &\cellcolor[HTML]{fbfada}0.539 &\cellcolor[HTML]{f5f4d9}0.752 &\cellcolor[HTML]{faf9d9}0.637 &\cellcolor[HTML]{fbfada}0.505 &\cellcolor[HTML]{f6f5d9}0.587 &\cellcolor[HTML]{e2e0d8}0.609 &\cellcolor[HTML]{eeecd9}0.671 &\cellcolor[HTML]{faf9d9}0.466 \\
Qwen-14B-Chat &\cellcolor[HTML]{f6f4d9}0.564 &\cellcolor[HTML]{f4f2d9}0.647 &\cellcolor[HTML]{f1f0d9}0.609 &\cellcolor[HTML]{f4f2d9}0.498 &\cellcolor[HTML]{f8f7d9}0.757 &\cellcolor[HTML]{fbfada}0.638 &\cellcolor[HTML]{f3f2d9}0.460 &\cellcolor[HTML]{fbfada}0.610 &\cellcolor[HTML]{efeed9}0.629 &\cellcolor[HTML]{f1f0d9}0.691 &\cellcolor[HTML]{fbfada}0.467 \\
Deepseek-LLM-67B-Chat &\cellcolor[HTML]{eeedd9}0.526 &\cellcolor[HTML]{ebe9d9}0.588 &\cellcolor[HTML]{f3f2d9}0.624 &\cellcolor[HTML]{ebe9d9}0.444 &\cellcolor[HTML]{d9d7d2}0.694 &\cellcolor[HTML]{edecd9}0.592 &\cellcolor[HTML]{e7e5d9}0.384 &\cellcolor[HTML]{f4f2d9}0.576 &\cellcolor[HTML]{e1dfd7}0.594 &\cellcolor[HTML]{edebd9}0.666 &\cellcolor[HTML]{f6f5d9}0.439 \\
GPT-3.5-Public-SFT &\cellcolor[HTML]{edecd9}0.522 &\cellcolor[HTML]{ebe9d9}0.587 &\cellcolor[HTML]{eae8d9}0.565 &\cellcolor[HTML]{f4f2d9}0.498 &\cellcolor[HTML]{f1f0d9}0.745 &\cellcolor[HTML]{ebe9d9}0.583 &\cellcolor[HTML]{f1efd9}0.444 &\cellcolor[HTML]{e5e3d9}0.501 &\cellcolor[HTML]{e7e5d9}0.620 &\cellcolor[HTML]{e9e7d9}0.643 &\cellcolor[HTML]{f8f7d9}0.452 \\
Yi-34B-Chat &\cellcolor[HTML]{ecebd9}0.516 &\cellcolor[HTML]{f1f0d9}0.631 &\cellcolor[HTML]{eeedd9}0.592 &\cellcolor[HTML]{edecd9}0.460 &\cellcolor[HTML]{fbfada}0.761 &\cellcolor[HTML]{e6e4d9}0.566 &\cellcolor[HTML]{f0efd9}0.440 &\cellcolor[HTML]{efedd9}0.551 &\cellcolor[HTML]{e2e0d8}0.610 &\cellcolor[HTML]{e2e0d8}0.608 &\cellcolor[HTML]{f1efd9}0.408 \\
Baichuan2-13B-Chat &\cellcolor[HTML]{ebead9}0.512 &\cellcolor[HTML]{edebd9}0.600 &\cellcolor[HTML]{eeedd9}0.591 &\cellcolor[HTML]{f0eed9}0.474 &\cellcolor[HTML]{f5f4d9}0.751 &\cellcolor[HTML]{efedd9}0.597 &\cellcolor[HTML]{efeed9}0.434 &\cellcolor[HTML]{e9e8d9}0.525 &\cellcolor[HTML]{dfddd6}0.572 &\cellcolor[HTML]{cdccc9}0.494 &\cellcolor[HTML]{eae9d9}0.372 \\
Tigerbot-13B-Chat &\cellcolor[HTML]{e8e6d9}0.494 &\cellcolor[HTML]{e9e7d9}0.571 &\cellcolor[HTML]{eae9d9}0.569 &\cellcolor[HTML]{e5e4d9}0.413 &\cellcolor[HTML]{eae8d9}0.732 &\cellcolor[HTML]{e4e2d9}0.560 &\cellcolor[HTML]{e4e2d9}0.365 &\cellcolor[HTML]{e5e3d9}0.502 &\cellcolor[HTML]{e2e0d8}0.607 &\cellcolor[HTML]{e1dfd7}0.601 &\cellcolor[HTML]{dedcd5}0.306 \\
Chinese-Alpaca-2-13B &\cellcolor[HTML]{e7e6d9}0.492 &\cellcolor[HTML]{e8e6d9}0.566 &\cellcolor[HTML]{e6e5d9}0.545 &\cellcolor[HTML]{e7e5d9}0.420 &\cellcolor[HTML]{e0ded7}0.712 &\cellcolor[HTML]{eeedd9}0.595 &\cellcolor[HTML]{e7e5d9}0.382 &\cellcolor[HTML]{e2e0d8}0.488 &\cellcolor[HTML]{fbfada}0.641 &\cellcolor[HTML]{fbfada}0.740 &\cellcolor[HTML]{e6e4d9}0.347 \\
Chinese-Alpaca-33B &\cellcolor[HTML]{e6e4d9}0.484 &\cellcolor[HTML]{e6e4d9}0.550 &\cellcolor[HTML]{dfddd6}0.506 &\cellcolor[HTML]{e7e5d9}0.423 &\cellcolor[HTML]{eae8d9}0.732 &\cellcolor[HTML]{e1dfd8}0.548 &\cellcolor[HTML]{dcdad4}0.342 &\cellcolor[HTML]{e3e1d9}0.494 &\cellcolor[HTML]{efeed9}0.629 &\cellcolor[HTML]{e9e8d9}0.648 &\cellcolor[HTML]{e4e2d9}0.334 \\
Ziya-Llama-13B &\cellcolor[HTML]{e5e3d9}0.479 &\cellcolor[HTML]{e5e3d9}0.546 &\cellcolor[HTML]{dddbd4}0.499 &\cellcolor[HTML]{e4e2d9}0.404 &\cellcolor[HTML]{f4f2d9}0.749 &\cellcolor[HTML]{eae9d9}0.582 &\cellcolor[HTML]{e4e2d9}0.367 &\cellcolor[HTML]{e4e2d9}0.499 &\cellcolor[HTML]{efeed9}0.629 &\cellcolor[HTML]{f7f6d9}0.722 &\cellcolor[HTML]{dfded6}0.313 \\
Chinese-Llama2-Linly-13B &\cellcolor[HTML]{e5e3d9}0.479 &\cellcolor[HTML]{e8e6d9}0.563 &\cellcolor[HTML]{e3e1d9}0.524 &\cellcolor[HTML]{e5e3d9}0.411 &\cellcolor[HTML]{d2d1cd}0.676 &\cellcolor[HTML]{e4e3d9}0.561 &\cellcolor[HTML]{e3e1d9}0.359 &\cellcolor[HTML]{e1dfd7}0.482 &\cellcolor[HTML]{e1dfd8}0.602 &\cellcolor[HTML]{f2f1d9}0.696 &\cellcolor[HTML]{e0ded6}0.313 \\
Tigerbot-7B-Chat &\cellcolor[HTML]{e5e3d9}0.478 &\cellcolor[HTML]{e3e1d9}0.532 &\cellcolor[HTML]{e8e6d9}0.554 &\cellcolor[HTML]{e2e0d8}0.393 &\cellcolor[HTML]{e9e8d9}0.731 &\cellcolor[HTML]{ebe9d9}0.583 &\cellcolor[HTML]{e0ded7}0.351 &\cellcolor[HTML]{e8e6d9}0.519 &\cellcolor[HTML]{f0efd9}0.630 &\cellcolor[HTML]{e3e1d9}0.614 &\cellcolor[HTML]{dad9d3}0.291 \\
ChatGLM3-6B &\cellcolor[HTML]{e3e1d9}0.472 &\cellcolor[HTML]{e7e5d9}0.557 &\cellcolor[HTML]{e3e2d9}0.526 &\cellcolor[HTML]{e3e1d9}0.397 &\cellcolor[HTML]{f4f2d9}0.749 &\cellcolor[HTML]{f3f2d9}0.612 &\cellcolor[HTML]{efedd9}0.431 &\cellcolor[HTML]{eae8d9}0.529 &\cellcolor[HTML]{e7e6d9}0.620 &\cellcolor[HTML]{deddd6}0.589 &\cellcolor[HTML]{eeecd9}0.392 \\
Chinese-Alpaca-13B &\cellcolor[HTML]{e3e1d9}0.471 &\cellcolor[HTML]{e1dfd8}0.524 &\cellcolor[HTML]{e0ded7}0.513 &\cellcolor[HTML]{e4e2d9}0.402 &\cellcolor[HTML]{e7e5d9}0.726 &\cellcolor[HTML]{dddbd5}0.526 &\cellcolor[HTML]{d1d0cc}0.323 &\cellcolor[HTML]{e2e0d8}0.486 &\cellcolor[HTML]{eeedd9}0.628 &\cellcolor[HTML]{f3f2d9}0.702 &\cellcolor[HTML]{e4e2d9}0.336 \\
ChatGLM2-6B &\cellcolor[HTML]{e1dfd8}0.464 &\cellcolor[HTML]{e0ded7}0.520 &\cellcolor[HTML]{e5e3d9}0.533 &\cellcolor[HTML]{e5e3d9}0.407 &\cellcolor[HTML]{e6e4d9}0.725 &\cellcolor[HTML]{d9d8d2}0.506 &\cellcolor[HTML]{e4e2d9}0.363 &\cellcolor[HTML]{e0ded7}0.480 &\cellcolor[HTML]{edebd9}0.627 &\cellcolor[HTML]{ecead9}0.661 &\cellcolor[HTML]{dddbd5}0.303 \\
Chinese-Alpaca-7B &\cellcolor[HTML]{dddbd4}0.452 &\cellcolor[HTML]{dbd9d3}0.504 &\cellcolor[HTML]{dddbd5}0.501 &\cellcolor[HTML]{d3d2ce}0.351 &\cellcolor[HTML]{dbd9d3}0.699 &\cellcolor[HTML]{e0ded7}0.543 &\cellcolor[HTML]{e4e2d9}0.365 &\cellcolor[HTML]{e0ded7}0.478 &\cellcolor[HTML]{eae8d9}0.623 &\cellcolor[HTML]{f5f4d9}0.711 &\cellcolor[HTML]{e3e1d9}0.328 \\
Chinese-Alpaca-2-7B &\cellcolor[HTML]{dbdad4}0.448 &\cellcolor[HTML]{dddbd5}0.509 &\cellcolor[HTML]{dbd9d3}0.493 &\cellcolor[HTML]{d1cfcc}0.344 &\cellcolor[HTML]{dcdbd4}0.703 &\cellcolor[HTML]{dad9d3}0.510 &\cellcolor[HTML]{d7d6d1}0.334 &\cellcolor[HTML]{e1dfd7}0.483 &\cellcolor[HTML]{f6f5d9}0.637 &\cellcolor[HTML]{e0ded7}0.596 &\cellcolor[HTML]{e6e4d9}0.343 \\
Chinese-Llama2-Linly-7B &\cellcolor[HTML]{d9d8d2}0.443 &\cellcolor[HTML]{d9d7d2}0.496 &\cellcolor[HTML]{e7e5d9}0.546 &\cellcolor[HTML]{d3d2ce}0.350 &\cellcolor[HTML]{e0ded7}0.713 &\cellcolor[HTML]{e4e2d9}0.559 &\cellcolor[HTML]{d2d0cd}0.323 &\cellcolor[HTML]{e3e1d9}0.495 &\cellcolor[HTML]{e1dfd8}0.603 &\cellcolor[HTML]{dddcd5}0.584 &\cellcolor[HTML]{dbd9d3}0.293 \\
Qwen-7B-Chat &\cellcolor[HTML]{d9d7d2}0.442 &\cellcolor[HTML]{d7d5d0}0.489 &\cellcolor[HTML]{e1dfd7}0.514 &\cellcolor[HTML]{dfddd6}0.384 &\cellcolor[HTML]{e0ded7}0.713 &\cellcolor[HTML]{e5e3d9}0.563 &\cellcolor[HTML]{d4d3ce}0.328 &\cellcolor[HTML]{dddbd5}0.463 &\cellcolor[HTML]{dfddd6}0.576 &\cellcolor[HTML]{e8e6d9}0.639 &\cellcolor[HTML]{d8d6d1}0.281 \\
ChatGLM-6B &\cellcolor[HTML]{d8d7d1}0.440 &\cellcolor[HTML]{d4d2ce}0.480 &\cellcolor[HTML]{d8d7d1}0.483 &\cellcolor[HTML]{d4d2ce}0.353 &\cellcolor[HTML]{edecd9}0.738 &\cellcolor[HTML]{d1cfcc}0.460 &\cellcolor[HTML]{dedcd5}0.346 &\cellcolor[HTML]{e0dfd7}0.480 &\cellcolor[HTML]{f3f2d9}0.633 &\cellcolor[HTML]{d6d4d0}0.543 &\cellcolor[HTML]{e2e0d8}0.322 \\
Baichuan-13B-Chat &\cellcolor[HTML]{d3d2ce}0.426 &\cellcolor[HTML]{e3e1d9}0.531 &\cellcolor[HTML]{edebd9}0.584 &\cellcolor[HTML]{cfcecb}0.339 &\cellcolor[HTML]{cfcecb}0.668 &\cellcolor[HTML]{d4d3ce}0.478 &\cellcolor[HTML]{eae8d9}0.402 &\cellcolor[HTML]{dcdbd4}0.459 &\cellcolor[HTML]{dddcd5}0.559 &\cellcolor[HTML]{cdccca}0.497 &\cellcolor[HTML]{eeecd9}0.392 \\
Yi-6B-Chat &\cellcolor[HTML]{d1d0cc}0.420 &\cellcolor[HTML]{d9d7d2}0.496 &\cellcolor[HTML]{e1dfd8}0.516 &\cellcolor[HTML]{d1cfcc}0.344 &\cellcolor[HTML]{f0eed9}0.742 &\cellcolor[HTML]{d6d5d0}0.488 &\cellcolor[HTML]{dfddd6}0.348 &\cellcolor[HTML]{e4e2d9}0.498 &\cellcolor[HTML]{edecd9}0.627 &\cellcolor[HTML]{d0cfcb}0.510 &\cellcolor[HTML]{d9d7d2}0.285 \\
CPM-Bee-10B &\cellcolor[HTML]{cfcecb}0.415 &\cellcolor[HTML]{cbcac8}0.451 &\cellcolor[HTML]{d5d4cf}0.472 &\cellcolor[HTML]{c3c2c2}0.304 &\cellcolor[HTML]{c7c6c5}0.647 &\cellcolor[HTML]{b8b7ba}0.329 &\cellcolor[HTML]{bdbcbe}0.284 &\cellcolor[HTML]{ecead9}0.538 &\cellcolor[HTML]{dbd9d3}0.534 &\cellcolor[HTML]{cbcac8}0.486 &\cellcolor[HTML]{dedcd5}0.305 \\
Moss-Moon-003-SFT &\cellcolor[HTML]{c9c8c7}0.399 &\cellcolor[HTML]{bcbcbd}0.403 &\cellcolor[HTML]{d1d0cc}0.457 &\cellcolor[HTML]{cac9c7}0.325 &\cellcolor[HTML]{e0ded6}0.712 &\cellcolor[HTML]{cfcecb}0.450 &\cellcolor[HTML]{c7c6c5}0.304 &\cellcolor[HTML]{d8d6d1}0.435 &\cellcolor[HTML]{e1dfd7}0.594 &\cellcolor[HTML]{d6d4d0}0.542 &\cellcolor[HTML]{deddd6}0.308 \\
Belle-SFT-Public &\cellcolor[HTML]{c9c8c7}0.397 &\cellcolor[HTML]{cacac8}0.450 &\cellcolor[HTML]{cac9c7}0.430 &\cellcolor[HTML]{cfcdcb}0.338 &\cellcolor[HTML]{c6c6c5}0.645 &\cellcolor[HTML]{cac9c7}0.426 &\cellcolor[HTML]{c5c4c4}0.300 &\cellcolor[HTML]{d1cfcc}0.398 &\cellcolor[HTML]{dddcd5}0.558 &\cellcolor[HTML]{f0eed9}0.683 &\cellcolor[HTML]{cac9c7}0.224 \\
Telechat-7B &\cellcolor[HTML]{b8b7ba}0.350 &\cellcolor[HTML]{b4b4b8}0.375 &\cellcolor[HTML]{c5c5c4}0.414 &\cellcolor[HTML]{b4b4b8}0.261 &\cellcolor[HTML]{cccbc9}0.660 &\cellcolor[HTML]{bababc}0.341 &\cellcolor[HTML]{cfcecb}0.320 &\cellcolor[HTML]{dddbd5}0.462 &\cellcolor[HTML]{f8f7d9}0.639 &\cellcolor[HTML]{cdccc9}0.494 &\cellcolor[HTML]{dddcd5}0.304 \\
Aquilachat-7B &\cellcolor[HTML]{b7b7ba}0.350 &\cellcolor[HTML]{b7b6ba}0.385 &\cellcolor[HTML]{c9c8c6}0.426 &\cellcolor[HTML]{b8b8bb}0.274 &\cellcolor[HTML]{b4b4b8}0.595 &\cellcolor[HTML]{b4b4b8}0.308 &\cellcolor[HTML]{b4b4b8}0.267 &\cellcolor[HTML]{d8d6d1}0.434 &\cellcolor[HTML]{e2e0d8}0.607 &\cellcolor[HTML]{bdbdbe}0.409 &\cellcolor[HTML]{e8e6d9}0.355 \\
Baichuan2-7B-Chat &\cellcolor[HTML]{b4b4b8}0.339 &\cellcolor[HTML]{bfbfc0}0.414 &\cellcolor[HTML]{b4b4b8}0.349 &\cellcolor[HTML]{cfcecb}0.339 &\cellcolor[HTML]{d1d0cc}0.673 &\cellcolor[HTML]{cbcac8}0.429 &\cellcolor[HTML]{c5c4c4}0.300 &\cellcolor[HTML]{b4b4b8}0.246 &\cellcolor[HTML]{b4b4b8}0.097 &\cellcolor[HTML]{b4b4b8}0.357 &\cellcolor[HTML]{b4b4b8}0.130 \\
\bottomrule
\end{tabular}
}
\end{table*}

\end{document}